%% file: sample-xelatex.tex
\documentclass[manuscript]{acmart}

\AtBeginDocument{%
  \providecommand\BibTeX{{%
    \normalfont B\kern-0.5em{\scshape i\kern-0.25em b}\kern-0.8em\TeX}}}

\usepackage{tabularx}
\usepackage{algorithm}
\usepackage{algpseudocode}

\usepackage[algo2e, ruled, vlined]{algorithm2e}

\theoremstyle{definition}

\begin{document}


\title{Large Language Models for User Interest Journeys}
\author{Konstantina Christakopoulou}
\author{Alberto Lalama}
\author{Cj Adams}
\author{Iris Qu}
\author{Yifat Amir} \author{Samer Chucri}
\author{Pierce Vollucci}
\author{Fabio Soldo}
\author{Dina Bseiso} \author{Sarah Scodel} 
\author{Lucas Dixon}
\author{Ed H. Chi}
\author{Minmin Chen}
\affiliation{%
  \institution{Google Inc.}
  \city{Mountain View}
  \country{USA}
}  

\renewcommand{\shortauthors}{Christakopoulou, Lalama, et al.}

\begin{abstract}
  \input{abstract}
\end{abstract}






\maketitle

\section{Introduction}
\input{introduction}
\label{sec:intro}

\section{Understanding Real User Journeys}
\label{sec:uxr}
\input{uxr}


\section{Methods: Journey Service}
\label{sec:journey_service}
\input{journey_service}

\subsection{Journey Extraction}
\label{sec:methods}
\input{methods}

\subsection{Journey Naming}
\label{sec:naming_journeys}
\input{method_naming}

\section{Journey Extraction Results}
\label{sec:results}
\input{results}
\section{Journey Naming Results}
\label{sec:results_naming}
\input{results_naming}


\section{Related Work}
\label{sec:formulation}
\input{formulation}

 \vspace{-0.1in}
\section{Conclusions and Future Work}
\label{sec:conclusions}
\input{conclusions}
\newpage
{
\footnotesize
\bibliographystyle{ACM-Reference-Format}
   \input{sample-xelatex.bbl}


}
\appendix

\end{document}

%% file: abstract.tex
Large language models (LLMs) have shown impressive capabilities in natural language understanding and generation. Their potential for deeper user understanding and improved personalized user experience on recommendation platforms is, however, largely untapped. This paper aims to address this gap.
Recommender systems today capture users' interests through encoding their historical activities on the platforms
. The generated user representations are hard to examine or interpret. On the other hand,  if we were to ask people about interests they pursue in their life, they might talk about their hobbies, like \emph{I just started learning the ukulele}, or their relaxation routines, e.g., \emph{I like to watch Saturday Night Live}, or \emph{I want to plant a vertical garden}. 
We argue, and demonstrate through extensive experiments, that LLMs as foundation models can reason through user activities, and describe their interests in nuanced and interesting ways, 
similar to how a human would. 

We define \emph{interest journeys} as the persistent and overarching user interests, in other words, the non-transient ones. These are the interests that we believe will benefit most from the nuanced and personalized descriptions. 
We introduce a framework in which we first perform \emph{personalized} extraction of interest journeys, and then summarize the extracted journeys via LLMs, using techniques like few-shot prompting, prompt-tuning and fine-tuning.  
Together, our results in prompting LLMs to name extracted user journeys in a large-scale industrial platform demonstrate great potential of these models in providing deeper, more interpretable, and controllable user understanding. We believe LLM powered user understanding can be a stepping stone to entirely new user experiences on recommendation platforms that are \emph{journey-aware}, assistive, and enabling frictionless conversation down the line. 

%% file: introduction.tex


With the abundance of the internet content, the role of recommendation systems has significantly expanded. 
In the past, users mainly relied on recommendation systems to make 
one-off decisions around where to eat, what to buy, or which movie to watch \cite{koren2009matrix, sarwar2001item, covington2016deep}. Nowadays, users expect the recommendation platforms to also support their persistent and overarching interests, including their real-life goals that last days, months or even years \cite{liang2023enabling, liang2019recommender, ekstrand2016behaviorism}. 
For example, a user who is into stand-up comedy would want recommendations tailored to their  tastes, and might expect the system to help them explore other forms of comedy performance. A user who is learning to play an instrument or is improving home decoration, would want recommendations and tips appropriate to their skill level. Or, a user with the goal of becoming an entrepreneur would want the recommender system to find inspirational content assisting them every step along the way. 

If one were to ask a friend for recommendations around any of their journeys, 
the friend would probably ask them to first describe their interests or needs in detail. 
Once they get a reply, like \emph{I want to know the history of stand-up comedy and the most famous stand-up comedian at this time}, or \emph{I started playing ukulele a month ago, and I want to improve my strumming skills}, 
the friend would then be in a much better position to give good recommendations. Conversely, recommender systems make recommendations by predicting the next item a user might want to interact with,  given their historical activities \cite{sarwar2001item, covington2016deep, beutel2018latent}. 

We argue that this type of collaborative filtering based approach \cite{he2017neural, covington2016deep} does not meet user needs for higher-level semantic preferences. In order for recommender systems to truly assist users through their real-life journeys, 
they need to be able to understand and reason about interests, needs, and goals users want to pursue \cite{radlinski2022natural, guha2015user, mehrotra2019jointly, sun2016contextual}. However, the task presents some challenges. First, users often do not explicitly spell out their interests, needs, and real-life goals to the recommenders. As a result, the recommenders need to infer them from the historical activities the users engaged on the platform. Second, users can have multiple journeys intertwined in their activity history at any time. Third and most importantly, journeys are personalized and nuanced. Two users who are both into stand-up comedy can be interested in completely different aspects of it (e.g.,  \emph{history of stand-up comedy documentaries} vs \emph{Saturday Night Live skits}). 
This is where Large Language Models (LLMs) come in play. LLMs have demonstrated impressive capabilities for natural language understanding and generation, achieving state-of-the-art performance in a variety of tasks, from coding to essay writing to question answering \cite{brown2020language, thoppilan2022lamda, chowdhery2022palm, chung2022scaling}. What if we power recommender systems with LLMs that can reason through user activities on the platform to uncover the underlying personalized and nuanced user interests, needs and goals, that is: user journeys?  

\begin{figure*}[!t]
    \centering
    \includegraphics[scale=0.35]{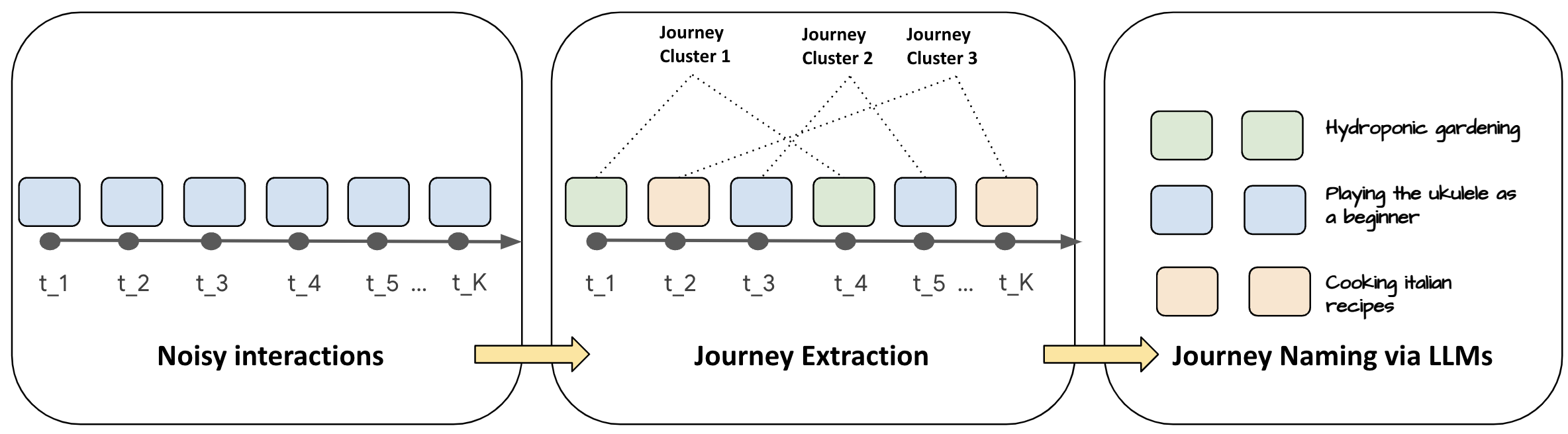}
    \caption{Our approach uses personalized clustering to uncover coherent user journeys, and names them via prompting LLMs.}
    \label{fig:journey_service}
\end{figure*}

To this end, we propose to build a personalized user journey profile which 1) uses personalized clustering to uncover coherent \emph{user journeys}, i.e., persisting user interests, needs, and goals, from a long sequence of user interaction logs, and 2) leverages the capabilities of LLMs aligned to the user interest journey domain through prompt-tuning \cite{lester2021power} and fine-tuning \cite{ziegler2020finetuning} on different data sources, to describe the extracted journeys with interpretable and nuanced names. Together, we make the following contributions:
\begin{itemize}
    \item A first demonstration of the capabilities of LLMs to uncover and describe in natural language the interests, needs, and goals users pursue, similar to how people would describe them, e.g., \emph{hydroponic gardening}, \emph{playing the ukulele as a beginner}, \emph{cooking italian recipes} (Figure \ref{fig:journey_service}). We posit that this will unlock unique user experiences and enable recommenders to assist users throughout their journeys.
    \item A thorough research study shedding light into the different factors impacting the quality of the generated journey names, e.g., the prompting techniques, the underlying domain data used for prompting, the LLM architecture and size, and the journey extraction technique. 
    \item An at-scale user research study uncovering the taxonomy of real users' journeys and how they pursue them on recommendation platforms.  
\end{itemize}

%% file: uxr.tex
To build recommender systems that are able to assist users on their journeys, we started by launching a series of user research studies consisting of online surveys and user interviews to understand the types of journeys users pursue. 
Informed consent from study participants was collected in accordance with company user data and privacy policies, which adhere to ethical research standards on human participants policies. Furthermore, all user data was de-identified by replacing names with randomized numbers and securely stored.

\noindent \textbf{Setup.} 
We sent 
online surveys to approximately $12,000$ users of an online platform, (a representative sample of the US internet user population aged 18-65+), with the goal of answering the questions: (Q1) If and how people use existing online platforms to pursue their journeys 
(Q2) What types of journeys people pursue online? 
Based on the analyzed survey data, we set up in-person interviews with a small cohort (N=9) of survey respondents who had multiple journeys to better understand: (Q3) What are the highlights and pain points of pursuing journeys on the internet? and (Q4) How do their journeys evolve? 
In the surveys, we framed journeys in terms of real-life interests or goals that the online platform has helped users pursue in a meaningful way. 


\begin{figure}
        \includegraphics[width=0.3\textwidth]{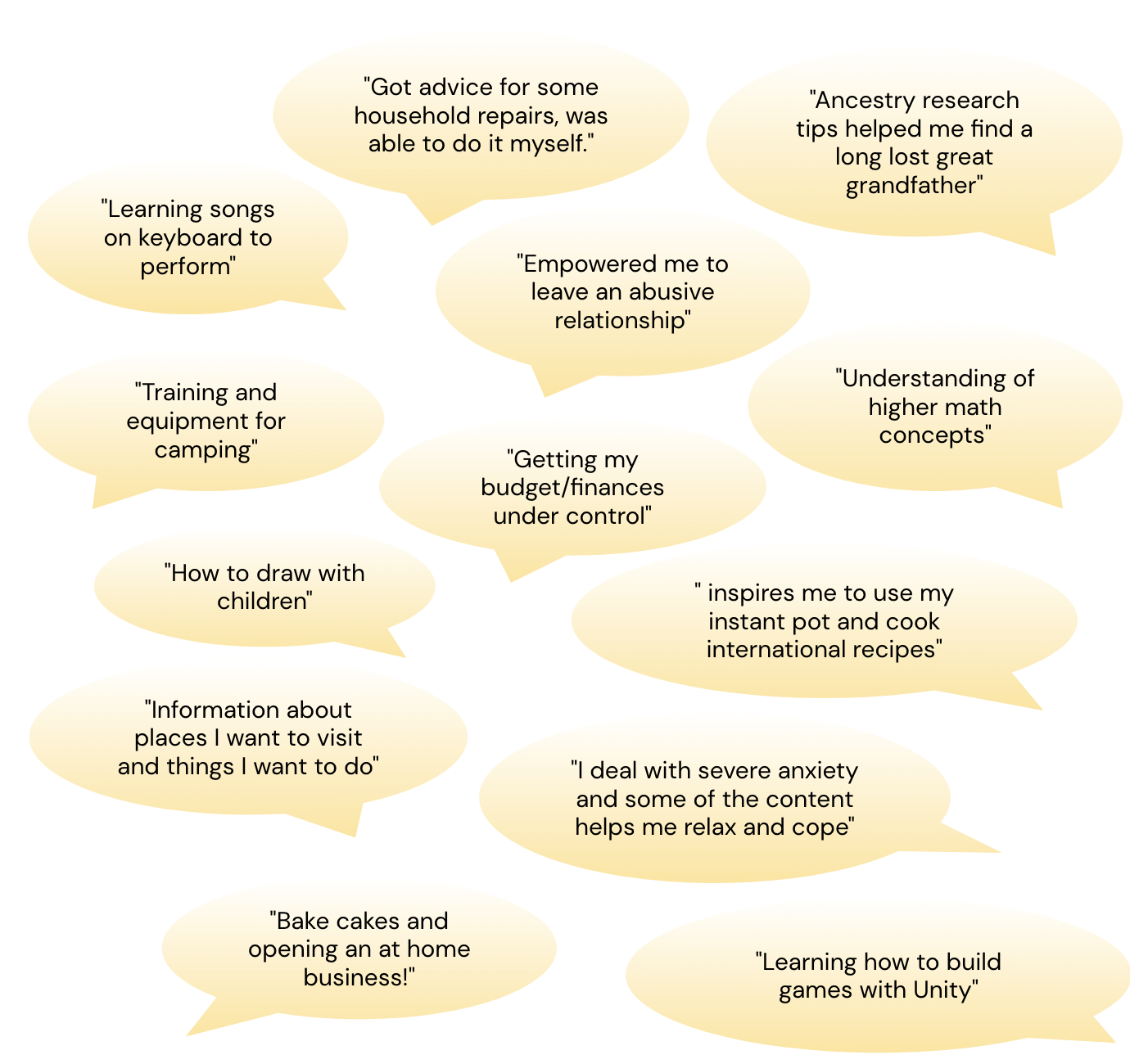}
       \includegraphics[width=0.3\textwidth]{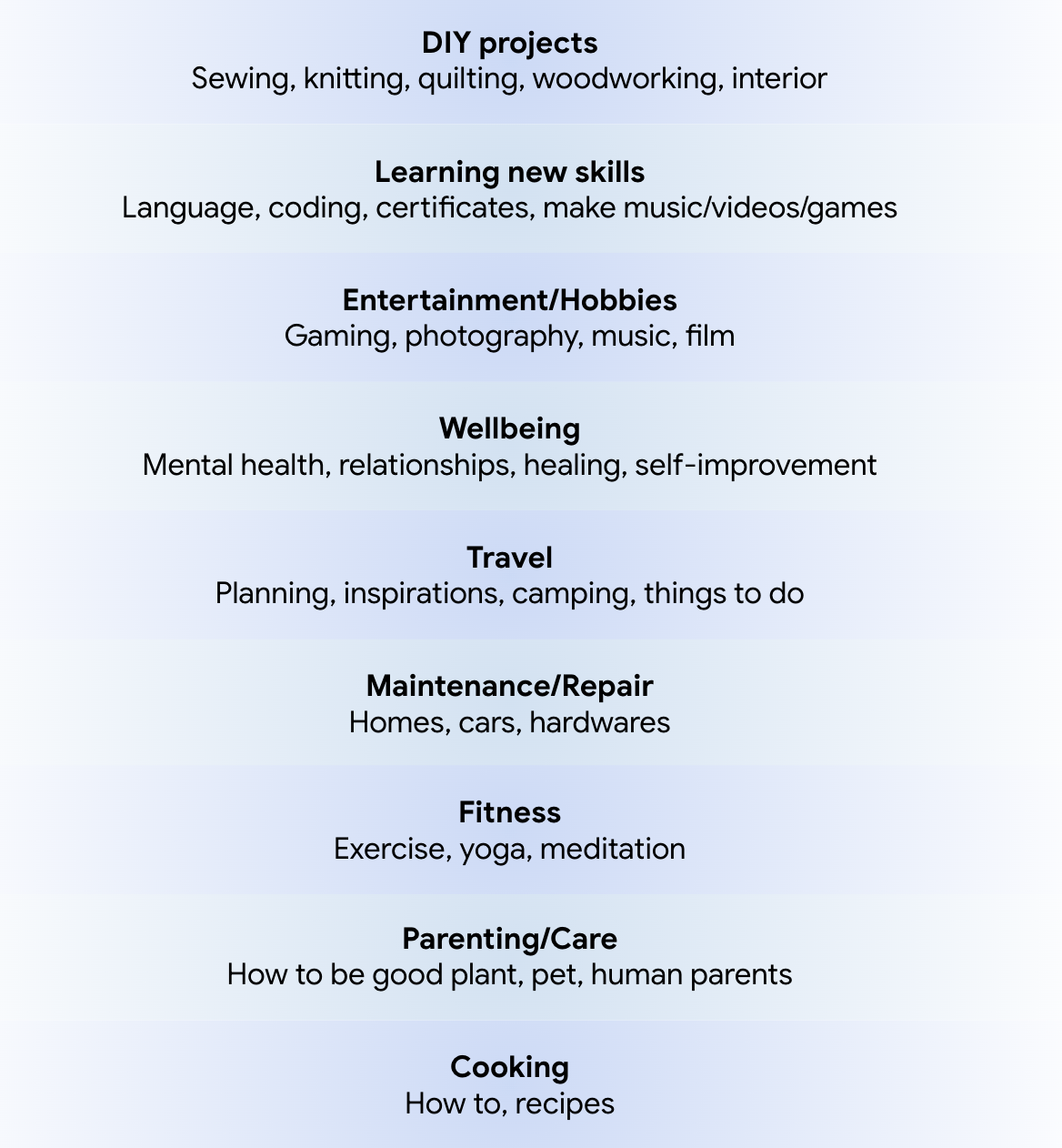}
    \caption{(left) Sample real journeys described by respondents. (right) Taxonomy of valued journeys uncovered by clustering text user responses via UMAP \cite{mcinnes2020umap}.
     }
      \label{fig:valued_journeys}
 \end{figure}
 
\noindent \textbf{Insight 1: \textbf{People value content related to entertainment, learning, and community engagement.}}
Figure \ref{fig:valued_journeys} (left) shows example real journeys as described by our respondents. Figure \ref{fig:valued_journeys}(right) shows the resulting taxonomy of valued interest journeys i.e., journeys they pursue and find satisfactory, by clustering the provided responses via UMAP \cite{mcinnes2020umap}, and identifying  roughly nine emerging themes. In a separate survey, we confirmed that entertainment (50 percent) and learning new skills (34 percent) are among the top valued journeys, 
and other dimensions include physical and mental well-being, caring for others, and community building etc. 

\noindent \textbf{Insight 2: \textbf{People pursue multiple interest journeys concurrently  
over long periods of time.}} 
Our surveys confirm that users do pursue real-life journeys on recommendation platforms. 
About $66\%$ of survey respondents used the platform recently to pursue a journey they valued. 
Out of the $66\%$, about 8 in 10 consumed content relevant to a journey for more than a month, with half saying some journey lasts for more than a year. 
People reported exploring multiple journeys simultaneously, with 7 in 10 pursuing one to three journeys and 3 in 10 exploring 4 or more in a single session. The large majority reported both exploring new valued interests and continuing existing ones. 

\noindent \textbf{Insight 3: \textbf{People rely more on explicit actions to find content relevant to their interest journeys.}} 
Half of our survey respondents picked the search bar as the most-used feature to pursue their journeys, compared to $20\%$ who relied on the recommendation feed, indicating inefficiency in existing recommenders in assisting user interest journeys, and the opportunity for providing more user control in recommendation experiences \cite{jannach2017user}. 


\noindent \textbf{Insight 4: \textbf{People's journeys are nuanced and evolve in a personalized way.}}
Our in-person interviews uncovered a lot of nuance in 
people's journeys. For example, one participant placed their interest journey in the general category of "gardening". As they dived deeper, they described it in much more nuanced phrases, i.e., "designing hydroponic systems for small spaces."  Another aspect uncovered was the right specificity people identify with, e.g.,   "greenhouse designs for cold climates" was deemed irrelevant for someone pursuing indoor gardening. 
Finally, participants mentioned how their interest can change based on offline and on-platform interactions. In other words, there are not two identical journeys between users, calling for personalized treatments. 


Together these research insights guide our methods and experimental setup, and provide motivation towards a recommendation experience that can offer  interpretability and control, enabling users to  pursue their interest journeys.

%% file: journey_service.tex


\subsection{Overview}
\label{sec:method:overview}

\textbf{Journeys.} 
We first define \emph{journey} as the umbrella term for user interests, needs, and goals: \emph{a user journey is a sequence of user-item interactions spanning different periods of time, coherent around a certain user interest, need or goal}. Users can have multiple journeys at any time point, and the journeys can be interleaving.

\noindent \textbf{Journey-aware recommendation.} We envision a journey-aware recommender system,  capable of identifying personalized user interest journeys and making recommendations accordingly to assist these journeys. 


Here we focus on the foundation work of extracting and naming user interest journeys, referred to as \textbf{journey service}, and leave mapping these extracted and named journeys to recommendations for future works. 
The proposed journey service consists of two components, as illustrated in Figure \ref{fig:journey_service}: 
\begin{enumerate}
    \item \textbf{Journey Extraction:} Maps the sequence of noisy interactions a user had with items on the platform into coherent journey clusters. 
    The journey clusters span different time periods, with singleton journeys (i.e., containing only a single user-item interaction) removed.
    \item \textbf{Journey Naming:} Maps the extracted journey clusters to human-readable, nuanced journey names based on content metadata, such as \emph{hydroponic gardening}, \emph{playing the ukulele as a beginner}. This enables users to comprehend how the system understands their journeys, and gives them control over their recommendations (e.g. they can choose more or fewer recommendations from a certain journey).
\end{enumerate} 
We study the necessity of these two components in section~\ref{sec:results} and 
\ref{sec:results_naming}.
Our user research uncovered properties we need to take into consideration when designing the journey service:

\noindent \textbf{(I) Granularity of journey clustering}. As revealed in the user surveys, journeys are personalized, thus defining the right granularity of journey clusters for all users is challenging. Measuring granularity quantitatively is hard, thus we propose to measure it via precision and recall, relying on proxy data (i.e., human-curated playlists) to align our journey clustering to how a person would define them (Section~\ref{subsec:analysis}).  

\noindent \textbf{(II) Nuance} \& \textbf{interestingness of journey names.} We would like to name the extracted journeys in a nuanced and interesting way, just as how the user would have described their own journey. For example, \emph{hydroponic gardening}, or \emph{growing indoor plants} as in Section \ref{sec:uxr}. We measure the quality of the generated journey names w.r.t. (proxy) ground-truth using Bleurt score \cite{sellam2020bleurt}, as well as imputed specificity, interestingness scores \cite{thoppilan2022lamda}, and other dimensions (Section~\ref{sec:results_naming}).  

\noindent \textbf{(III) Safety of journey names.} As we envision showing users the generated names and allowing them to control and refine their recommendations down the line, it is important that the generated names are safe \cite{thoppilan2022lamda}. We rely on imputed safety score to measure safety of the generated names. We find the safety of the journey name largely depends on the safety of the included items within the journey cluster (Section~\ref{sec:results_naming}). 

%% file: methods.tex
The journey extraction component maps a user's overall history into coherent journey clusters. 
The most straight-forward approach is to partition the entire item corpus into clusters, where each cluster represents one journey cluster. The caveat is that a single item can often cover multiple nuanced journeys. For example, an item on ``cooking with herbs for improved mental clarity'' can be a part of a ``Mediterranean cooking journey", ``improve your gut health" journey or ``grow your herbs garden journey". A global clustering would then mix these more nuanced journeys together. 
As suggested by our user research, although there are commonalities among user journeys, there are no two identical journeys across users. We thus hypothesize that
\emph{a personalized view of journeys is more appropriate than assuming space of journeys is global}, and propose an infinite concept personalized clustering algorithm.
\label{subsec:personalized_clustering}

\begin{figure}
\begin{minipage}{0.48\textwidth}
\begin{algorithm}[H]
\SetAlgoLined
\begin{singlespace}
\caption{Infinite Concept Personalized Clustering (ICPC) on a User}
\label{algorithm:sts}
\begin{algorithmic}
\small
\Statex \textbf{Input:} ${\mathcal H}_u = \{i_t, t=1, \cdots, T\}$ containing list of items the user interacted with; $\epsilon \in [0, 1]$: salient terms similarity threshold; $c$: Minimum number of items per cluster. Default values: $\epsilon=0.1, c=1$. 
\State \textbf{Initialize:} Journey Set ${\mathcal S}_J = \emptyset$ 
\State \For{$t = 1, \cdots, T$} {
$\forall J \in {\mathcal S}_J$, compute ItemJourneySim($i_t, J$)  \\
$J^\ast$ $\gets$ $\arg\max_{J'\in {\mathcal J}}$ ItemJourneySim($i_t, J'$) \\
\uIf{ItemJourneySim($J^\ast$, $i$) $\ge \epsilon$} {
$J^\ast = J^\ast \cup \{i_t\}$
} 
\Else{
Start a new journey $J_{\text{new}} = \{i_t\}$, ${\mathcal S}_J = {\mathcal S}_J\cup \{J_{\text{new}}\}$. 
}
Update the journey representation based on the added item.\\
}
\State Prune journey clusters with less than $c$ items 
\end{algorithmic}
\end{singlespace}
\end{algorithm}
\end{minipage}
\quad\quad
\begin{minipage}{.45\textwidth}
\vspace{-0.1in}
        \includegraphics[width=\textwidth]{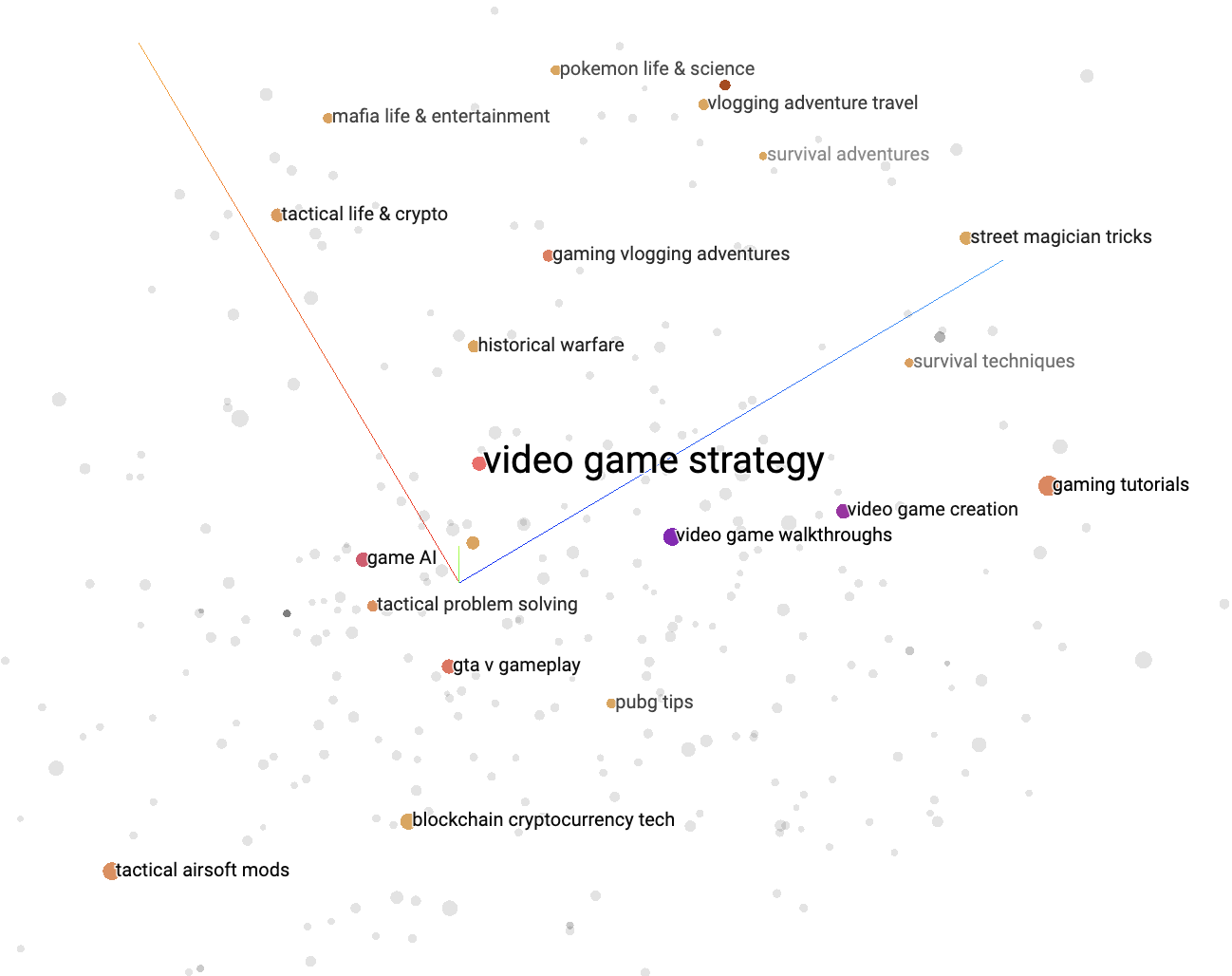}
        \caption{Visualization of journeys across users, with each point representing an extracted and named journey. We generate embeddings for each journey name through Universal Sentence Encoder \cite{cer2018universal} and cluster them with UMAP \cite{mcinnes2020umap}. Highlighted, a \emph{video game strategy} journey and its nearest neighbors in the embedding space, per cosine similarity.
        }
        \label{fig:journey_visualization}
    \end{minipage}%
\end{figure}


\textbf{Infinite concepts.} We start with annotating each item with a set of  salient terms (unigrams and bi-grams), each with an associated weight called salience score \cite{37686, gamon2016identifying}. Sources for the unigrams and bigrams are the text (e.g., title, description) of the item, search query that lead to clicks on the item, anchor text, and so on. Together these terms provide the semantic meaning of the item. The salience score of each term is always between $[0, 1]$, which is obtained by training a supervised learning model to predict human ratings of relevance between the salient term and the item, with tf/idf demotions built in. One can then obtain the salient term representation of a group of items by aggregating salient terms of individual items, taking into account their salience score, resulting in a new set of salient terms representing the item group. This way, any group of items (or any concept for that matter) can be represented in the salient term space, which can be thought of as an infinite dimension embedding space where one salient term represents one dimension. Furthermore, the similarity between any two concepts can be computed in terms of cosine similarity of the salient terms embeddings, measuring the overlap of salient terms.

\textbf{Infinite Concept Personalized clustering (ICPC).} As depicted in Algorithm 1
, our proposed algorithm effectively performs online clustering of a user's interaction history, based on the salient terms of the comprising items. Each user starts with an empty journey cluster set. We parse the user historical interactions one by one, each time assigning an item to the nearest, in terms of salient terms cosine similarity, journey cluster. Meanwhile, the representation of the assigned cluster is updated with the salient terms and salience scores of the newly assigned item. If no cluster is nearby (determined by thresholding the cosine similarity), a new cluster is created with the item assigned to it. The algorithm outputs the  resulting journey clusters, which by definition will be thematically coherent, as shown in figure~\ref{fig:journey_visualization}. The highlighted journey, extracted and named as ``video game strategy" is closest to journeys such as ``video game walkthroughs" and ``game AI". 

%% file: method_naming.tex
Once we have extracted journey clusters for a user, we describe each journey in natural language so that it is explainable, scrutable, and controllable \cite{balog2019transparent}. 
This is where Large Language Models (LLMs) come into play. Here, we describe the underlying models we leverage (Section \ref{subsec:models}) and how we adapt them to the domain of user interest journeys (Section \ref{subsec:alignllms}) to capture nuanced natural language descriptions.

\subsubsection{Models}
\label{subsec:models}
In this paper, we build on LaMDa \cite{thoppilan2022lamda} and PaLM \cite{chowdhery2022palm} family of LLMs.

\textbf{LaMDA} \cite{thoppilan2022lamda}, aka. Language Models for Dialog Applications, is a family of Transformer-based, decoder-only, neural language models specialized for dialog, which have up to 137B parameters and are pre-trained on 1.56T words of public dialog data and web text. The models have shown state of the art performance across tasks, including the BigBENCH \cite{srivastava2022imitation}, and have been fine-tuned on human rewrites so to provide safe, interesting, sensible, and specific dialog responses. 

 \textbf{PaLM} \cite{chowdhery2022palm}, aka. Pathways Language Model, is a densely-activated decoder-only transformer language model trained using Pathways \cite{barham2022pathways}, a large-scale ML accelerator orchestration system that enables highly efficient training across TPU pods. At the time of release, PaLM 540B achieved breakthrough performance on a suite of multi-step reasoning tasks \cite{chowdhery2022palm}. Its training corpus consists of 780 billion tokens representing a mixture of webpages, Wikipedia articles, source code, social media conversations, news articles and books \cite{chowdhery2022palm}. 
 
Although we mainly build on LaMDA and PaLM, in our experiments we also considered \textbf{Flan-PaLM}, i.e., the instruction-tuned counterpart introduced by Wei et al.\cite{wei2022finetuned}. This pre-trained model was fine-tuned on datasets where each example is prefixed with some combination of instructions and/or few-shot exemplars, and was shown in \cite{chung2022scaling} to outperform PaLM on several benchmarks. We also considered \textbf{PaLMChilla}, i.e., a PaLM variant trained on an additional data mixture that follows the Chinchilla compute-optimal training procedure \cite{hoffmann2022training} approach.



\subsubsection{Prompting LLMs for user interest journeys}
\label{subsec:alignllms}
Although general-purpose LLMs have reached state of the art performance on a wide variety of tasks on challenging benchmarks \cite{wang2020superglue}, for them to have good performance for the domain of user journey understanding, it is necessary to align them using domain-specific data. 
We explore data-efficient techniques of \textbf{few-shot prompting} \cite{brown2020language} and \textbf{prompt-tuning} \cite{lester2021power}, and data-rich \textbf{end-to-end fine-tuning} \cite{ziegler2020finetuning}.

\textbf{Few-shot prompting} 
Brown et al. \cite{brown2020language} demonstrated that LLMs are strong few-shot learners. Fast in-context learning can be achieved through including a handful of demonstration examples as prompts in the input context, \emph{without any gradient updates or fine-tuning}. In this study we focused on standard zero-shot, and few-shot. 
\emph{Zero-shot prompting} queries these models with an instruction describing the task without any additional examples, i.e., \emph{Summarize my interest journey in a concise and interesting way}. 
In \emph{few-shot prompting}, the prompt further includes few-shot examples describing the task through text-based demonstrations. These demonstrations are encoded as input-output pairs. 

\textbf{Prompt tuning} \cite{lester2021power} has been proposed as a data efficient domain adaption technique for LLMs. The key idea is that instead of changing \emph{all} model parameters, only a small subset of parameters corresponding to the prompt embedding will be updated given a small training set of input-output pairs (in the order of tens to hundreds). This can be viewed as a soft prompt, as opposed to the hard prompt encoded in few-shot prompting. 

\textbf{Fine-tuning} \cite{ziegler2020finetuning} updates \emph{all} model parameters to adapt to the new domain given a large amount of in-domain data. 
Typically, fine-tuning can achieve the best in-domain results with in-domain data in the order of thousands, but is also the most costly in compute and maintenance. An advantage of prompt-tuning compared with fine-tuning, is that the same underlying model (with  same parameters) can be adapted to different domains by using prompt embeddings tuned on the domain data.
 
 To be consistent across prompting strategies,  we pre-processed input-output examples to follow the same format, as shown in Table \ref{tab:format_examples}. 


\begin{table}
\small{
    \centering
    \newcolumntype{s}{>{\hsize=.5\hsize}X}
    \newcommand{\heading}[1]{\multicolumn{1}{c}{#1}}
    \renewcommand*{\arraystretch}{1.2}
    \begin{tabularx}{\textwidth}{XXs} 
    \toprule
        Prompt Template & Input & Output \\
        \midrule
        I consumed content with titles: \{ \}.\newline I would describe one of my interests as: &
        Natural \& Cultural Heritage Conservation in the Philippine Seas; Seafaring Vessels of Ancient Filipinos; 
        The Philippines as a Maritime Nation: Opportunities and Challenges &
        Philippine \newline Maritime History \\
        \hline
        titles: \{ \} interest\_journey: &
        Two-Angle Bisectors; Equal Angles; Cyclic Quad; Equal Segment; Circumcircles &
        Olympiad Geometry \\ \hline
        keywords: \{ \} interest\_journey: &
        history, microwave, nerd, nostalgia, mp3 
        & Evolution of Personal Technology \\
        \bottomrule
    \end{tabularx}
    \caption{Prompt formats: (a) natural language titles prompt, used across all prompting strategies unless stated otherwise; (b) structured titles prompt; (c) structured playlist infinite concepts prompt.
    }
    \label{tab:format_examples}
    }
\end{table}


 \subsubsection{Data for aligning LLMs to user journeys}
 \label{subsec:data_alignment}
 To align LLMs to a new domain, it is important to have high-quality examples from that domain. Different prompting techniques, as discussed above, require different volumes of data. 
The ideal dataset for our task is derived from asking users to annotate and describe interest journeys they are pursuing on the platform in a nuanced and interesting way. This is, however, difficult and time-consuming to achieve at scale. 

\textbf{Fine-tuning data.} To overcome the challenge, we rely on user-curated playlists on the platform as proxy datasets. Each playlist maps groups of content about the same topic to a user-provided name. We only include playlists that have high episodicity (items tend to be consumed in a sequential order) and are classified as learning-focused by a neural network  trained on raters data. We refer to this dataset as \textbf{learning playlists}. The upside is that there are a lot of learning playlists readily available to fine-tune the model (we used twenty thousands examples); the downside is that the ground-truth playlist names tend to be short and noisy, without necessarily the nuanced nature we are looking for.

\textbf{Prompt tuning data.} Prompt tuning requires much fewer examples compared to fine-tuning. High-quality data, however, is crucial to ensure accurate results. Thus, instead of relying on the noisy learning playlists, we curated three small data sources for prompt tuning our models: (D1) \textbf{user interviews}, (D2) \textbf{user collections}, (D3) \textbf{expert-curated collections}, all in the order of less than a hundred examples. The (D1) source is the closest approximation to the ideal dataset, which is collected during our user interviews. We asked 50 users to describe journeys they pursued on the platform for more than a month, and pick from their interaction history the items that helped them toward these journeys. 
Source (D2) uses data from a separate platform anchored towards long-term aspirational journeys, where users save items in collections 
that they annotate with collection names.  
Lastly, (D3) comes from learning editorial collections, where expert editors have curated collections of items tailored to learning topics, and have named them in a nuanced and interesting way.

\textbf{Few-shot examples} For few-shot prompting, we only need a handful of examples (five to ten) from the domain of user interest journeys. For consistency, we sample examples from one of the above mentioned prompt tuning data.

%% file: results.tex


\subsection{Experimental Setup}
\label{subsec:analysis}

\begin{figure}[!t]
    \centering
    \includegraphics[width=0.95\textwidth]{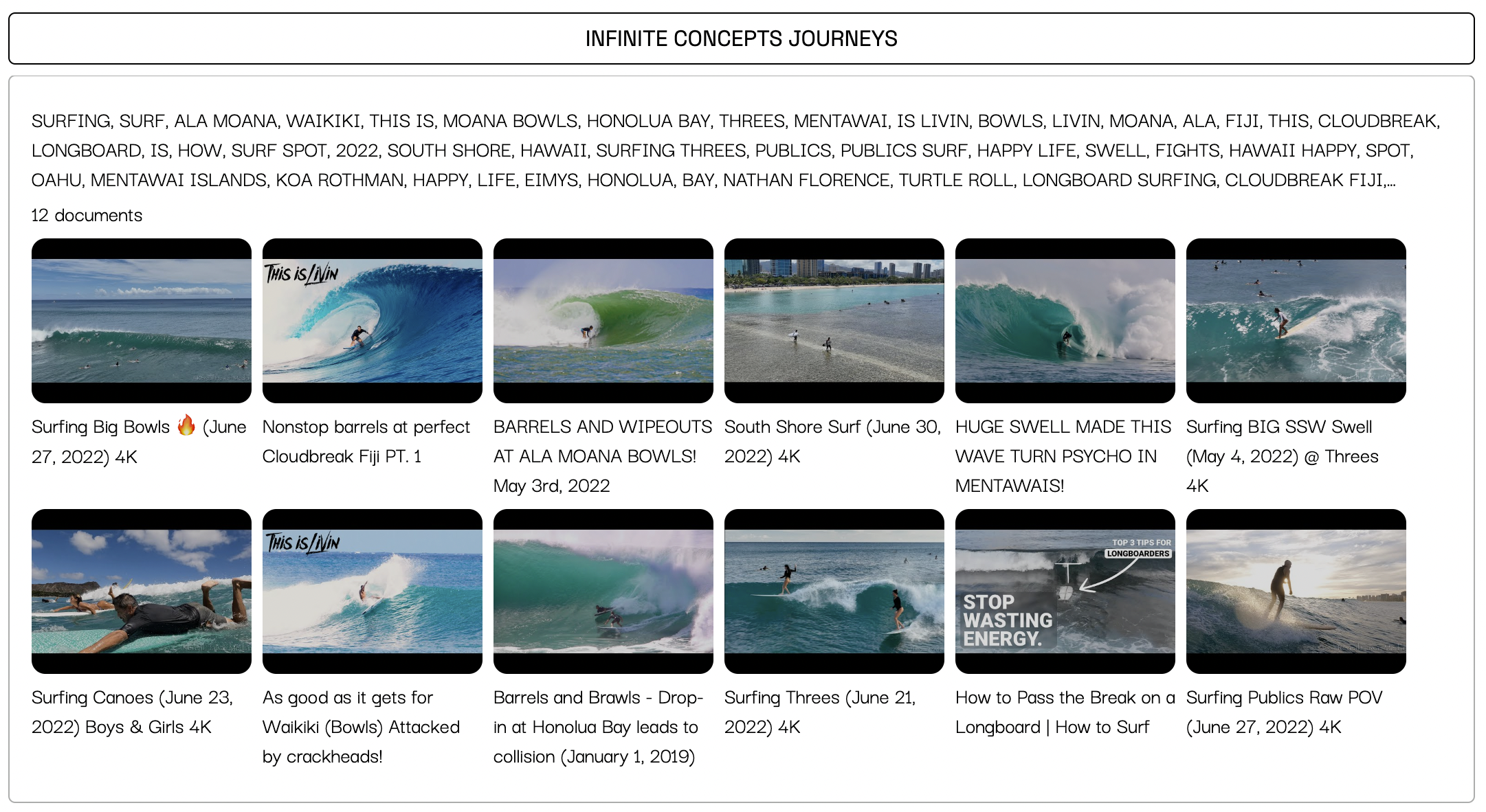}
    \begin{minipage}{\textwidth}
    \centering
    \includegraphics[width=0.475\textwidth]{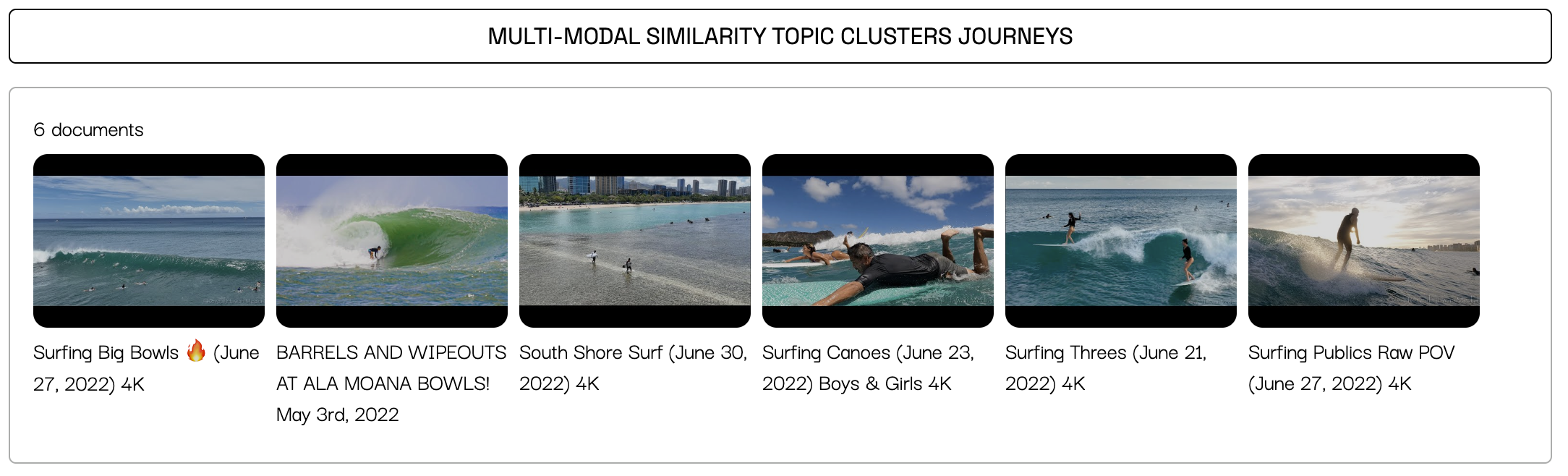}
    \includegraphics[width=0.475\textwidth]{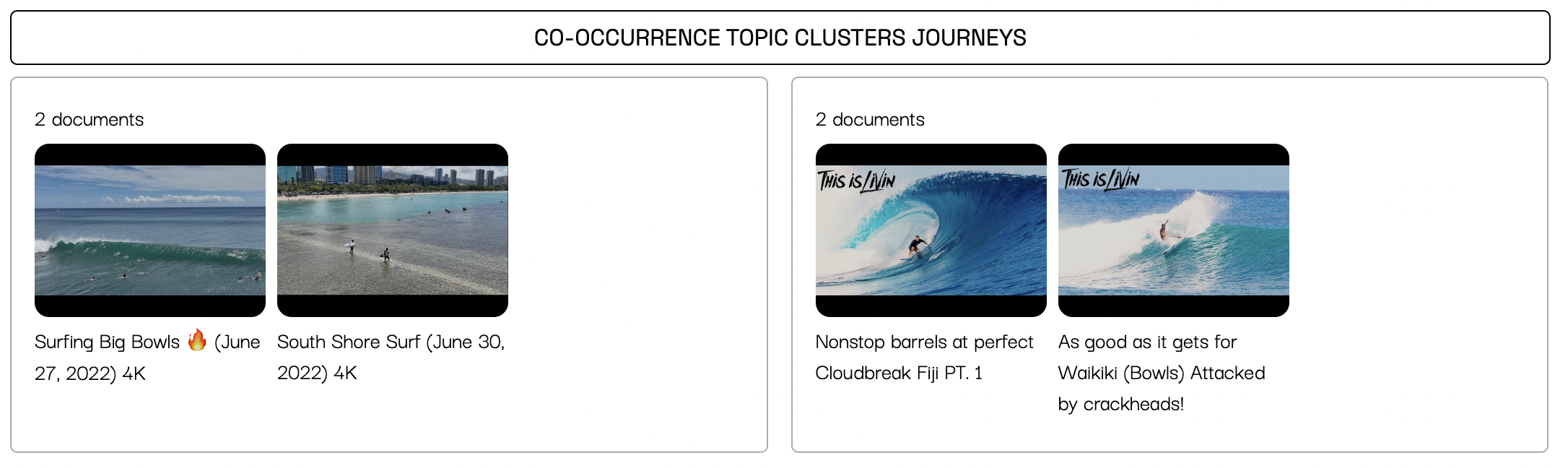}
    \end{minipage}
    \hfill
    \caption{Visual depiction of an infinite concepts personalized clusters journey extraction for \emph{surfing}. Shown at the top, the journey's salient terms representation. Below, the set of documents with a thumbnail and title for each. In contrast, multi-modal similarity topic clusters journeys only retrieve 6 documents, and co-occurrence topic clusters split these into 2 clusters, each with 2 documents.}
  \label{fig:visual_journeys}
  \vspace{-0.2in}
\end{figure}

\textbf{Metrics.} As discussed in Section~\ref{sec:method:overview}, measuring the right granularity of the extracted journey is extremely challenging. We look at qualitative results when extracting journeys on real user histories (evaluation data E1 explained below). Figure \ref{fig:visual_journeys} shows one comparison between journeys extracted using different methods. To obtain quantitative results comparing different journey extraction algorithms, we measure \emph{precision} (percentage of items correctly assigned the right journey cluster) and \emph{recall} (percentage of items belonging to each journey cluster getting retrieved) on a ``golden journey'' set detailed in the following (evaluation data E2). 

\textbf{Evaluation Data} We created two evaluation sets to evaluate journey extraction: 
\begin{enumerate}
    \item[E1] \textbf{Unlabeled Histories:} 300 user interaction histories, each with at least 10 valued interactions over 30 days, resulting in a dataset of 18,370 user-item interactions. 
    We define a valued user-item interaction as one where 1) the user engaged with at least $X$ minutes, and 2) a model that predicts user satisfaction surveys has found that the predicted user satisfaction score is in the top percentiles. The 300 user histories have a median length of 31.5, a mean length of 61.23 and a standard deviation of 69.41. The longest one contains 479 items; the shortest, 11.
    \item[E2] \textbf{Learning playlists as golden journeys:} 
    We sampled 5,000 learning playlists as explained in Section~\ref{subsec:data_alignment} as the golden journeys, comprised in total of 130,381 items. For each user we randomly sample and mix two playlists/journeys, and the goal is to cluster the items back into the two playlists.
\end{enumerate}

\textbf{Baselines.} We compared with baseline approaches that uses different strategies to partition the entire item corpus into non-personalized global clusters; each cluster then becomes one journey cluster: 

\begin{enumerate}
    \item \textbf{Clusters based on co-occurence behavior}. A topic cluster for each item is produced by: 1) taking the item co-occurrence matrix, where entry (i, j) counts the number of times item i and j were interacted with by the same user consecutively; 2) performing matrix factorization to generate one
embedding for each item; 3) using k-means to cluster the learned embeddings into $K$ clusters; 4) assigning each item to the nearest cluster centroid.
    \item \textbf{Clusters based on multimodal item similarity} These clusters group items together based on their audiovisual similarity, using online agglomerative/hierarchical clustering. The first-level of the hierarchy is capturing the macro-clusters, while the second level contains items all of fixed distance $\epsilon$; each macro cluster would then serve as a journey cluster. Online clustering works by comparing each new item with the micro-clusters in terms of embedding similarity, and assigns it to the closest micro-cluster. If too often an item is found in close proximity to multiple micro-clusters, micro-clusters are merged.
\end{enumerate}

\textbf{Parameters.} For the proposed Infinite Concept Personalized Clustering (we will refer to it here as \emph{ICPC} for brevity), we set the similarity threshold parameter to 0.1 unless explicitly stated otherwise, as we found qualitatively that this results in the most coherent journeys of the right granularity. For the baselines, we used parameters as tuned in a large-scale production recommender system. For the co-occurrence based topic clusters, the total number of clusters $K$ is 10,000; for multi-modal similarity-based clusters, the distance $\epsilon$ is set to 70. For all methods, we prune resulting clusters with a single item (we specify when we prune those with less than 5 items). 



\subsection{Key Results}
\begin{figure}
\vspace{-0.3in}
\includegraphics[width=0.45\textwidth]{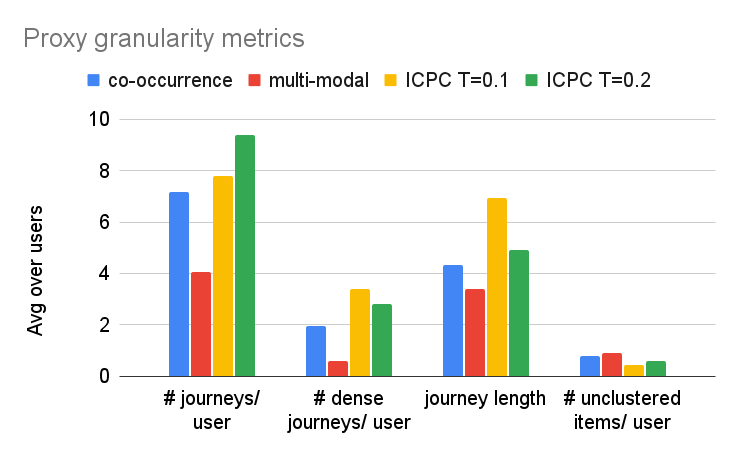}
\caption{\textbf{Comparison of journey extraction methods across proxy granularity metrics in (E1) setup.}}
\label{fig:extraction_stats_per_user}
\vspace{-0.1in}
\end{figure}Here we qualitatively and quantitatively compare the extracted journeys using our proposed ICPC algorithm and baselines across the two experimental setups.

\subsubsection*{ICPC achieves long coherent journeys, and is more robust to trivial coherence as obtained by singleton journey clusters.}
\label{subsec:docs_per_journey}
As shown in Figure~\ref{fig:visual_journeys} and Figure \ref{fig:extraction_stats_per_user}, the proposed method produces longer coherent journeys than baselines. 
The baselines on the other hand tend to produce trivially coherent journeys by assigning each item to a different cluster. 
Figure \ref{fig:extraction_stats_per_user} shows that a fraction of 0.79 of items per user are singletons for the co-occurrence based baseline; the number increases to 0.90 for the multi-modal similarity-based ones. This fraction is much lower, but exists for our method too ($\sim0.4$).

\subsubsection*{ICPC 
corroborates our user research findings that each user pursues a few journeys.}
    As shown in 
    Figure \ref{fig:extraction_stats_per_user}
    , restricting the minimum number of items per journey to a value of 5, i.e., considering only dense journeys, the mean number of journeys per user to significantly decrease. The decrease is very prominent for multi-modal similarity topic clusters, suggesting that clusters are sparser. Conversely, \emph{the cardinality of ICPC clusters and co-occurence topic clusters is consistent with user research} (Section \ref{sec:uxr}), that 7 of out 10 users pursue 1-3 interests simultaneously and 3 of out 10 sustain 4 or more.

\begin{figure}
\centering
    \small{
    \begin{tabular}{ | c || c | c | c |}
          \hline
           & \multicolumn{3}{|c|}{Two golden journeys per user} \\
           \hline \hline
           \textbf{Method} & \textbf{Recall} & \textbf{\# journeys} & \textbf{\# clusters / journey} \\ \hline
           Co-occurrence & 0.29 & 106,936 & 4.56 \\ \hline
           Multi-modal & 0.16 & 71,388 & 3.63 \\ \hline
           ICPC (\textbf{ours}) & \textbf{0.82} & \textbf{13,488} & \textbf{2.42} \\ 
           \hline
    \end{tabular}
    }

    \caption{\textbf{Comparison of journey extraction methods in E2.}}
    \label{tab:extraction_stats_per_user}
\end{figure}

\subsubsection*{ICPC 
can easily control journey granularity via the similarity threshold.}
We find that at a lower threshold, more items are grouped into the same journey cluster, resulting in a coarser granularity. Increasing the threshold renders a larger number of journeys that are more specific, unearthing deeper semantic relationships between items. For example, in a public transportation-related example journey, we find that a 0.1 similarity threshold breaks up the \textit{public transportation} journey (with 0.05 threshold) into three journeys: a general \textit{public transportation} journey, a more specific \textit{London public transportation} and a \textit{public transportation video game simulations}. With pre-defined global clusters, it is difficult for the baselines to produce journey clusters at different granularity.

\subsubsection*{ICPC achieves the highest recall.}
Considering the (E2) setup of two golden journeys/playlists per user, we see from Figure \ref{tab:extraction_stats_per_user} that our ICPC algorithm 
 achieves much higher recall compared with baselines.
On average, 2.42 journey clusters are extracted per user using ICPC; close to the ground truth of 2 journeys per user.  
From the total \# of journeys, we see that our method retrieves 1.3 more journey clusters than golden, versus co-occurrence gets around 10.7 more journey clusters. 

Together, our findings suggest that our proposed ICPC approach can effectively cluster noisy interaction histories into journeys with the desired properties, significantly outperforming the non-personalized baselines. 

%% file: results_naming.tex
Next, we evaluate the quality of journey naming in describing user interest journeys. We provide extensive analysis to understand the effect of different prompting techniques, underlying data, and models. 

\subsection{Experiment setup} 
We evaluate journey naming on  three experimental setups.
\begin{enumerate}
    \item[N1] \textbf{Expert-curated collections}: Data of 100 item collections, named with interesting engaging names from editorial teams. The set comes from the same distribution as (D3) used for prompt-tuning (Section \ref{subsec:data_alignment}), but it is from a different unseen split.
    \item[N2] \textbf{Learning Playlists}: Larger-scale 10,000 playlists, of high episodicity focused in learning-related goals. The set comes from the same distribution as the learning playlists considered for fine-tuning (Section \ref{subsec:data_alignment}); again a different unseen split. As explained before, the names are shorter and noisier, 
    but still convey meaning about the items saved in the playlists.
    \item[N3] \textbf{Unlabeled Histories}: Larger scale 10,000 unlabeled extracted journeys, from histories of 2,000  users spanning a month (same distribution as E1 in Section \ref{subsec:analysis}). The journeys have been extracted using our ICPC algorithm. These journeys do not have expert or user-associated names. 
\end{enumerate}


For each journey, we concatenate titles of all the items belonging to the journey as input context to the LLMs as shown in Table~\ref{tab:format_examples}, and generate the journey name using different LLMs with different prompting techniques. Unless otherwise specified, we mainly build on top of the LaMDA-137B foundation model \cite{thoppilan2022lamda}. We set the prompt embedding size to 5 tokens. We perform 10,000 prompt tuning steps, and 20,000 fine tuning steps. The learning rate and dropout rate is set to 3e-5 and 0.1 respectively across prompt and fine-tuning. During inference, we set beam size to 1, temperature to 0, max decoding steps to 256 and feature lengths 256 tokens for the input and 64 tokens for the targets.

To quantitatively compare the different naming user journeys approaches, we  rely mainly on BLEURT \cite{sellam2020bleurt} and SacreBLEU \cite{papineni2002bleu} scores.
While SacreBLEU measures the  overlap between the generated and the ground truth name, BLEURT, as an embedding-based metric, measures their semantic similarity. For the (N3) setup where we do not have the ground truth names, we propose to rely on a BLEURT score between the concatenated names of input items comprising the user journey and the generated name as the metric.\footnote{BLEU  has been developed and can be interpreted for the domain of machine translation; thus the interpretation of the values for our domain does not have a known reference point. That's why we use it for comparison rather than absolute scores, and we rely on qualitative evaluation as well.}  

\begin{figure}[!th]
    \centering
    \includegraphics[width=0.9\textwidth]{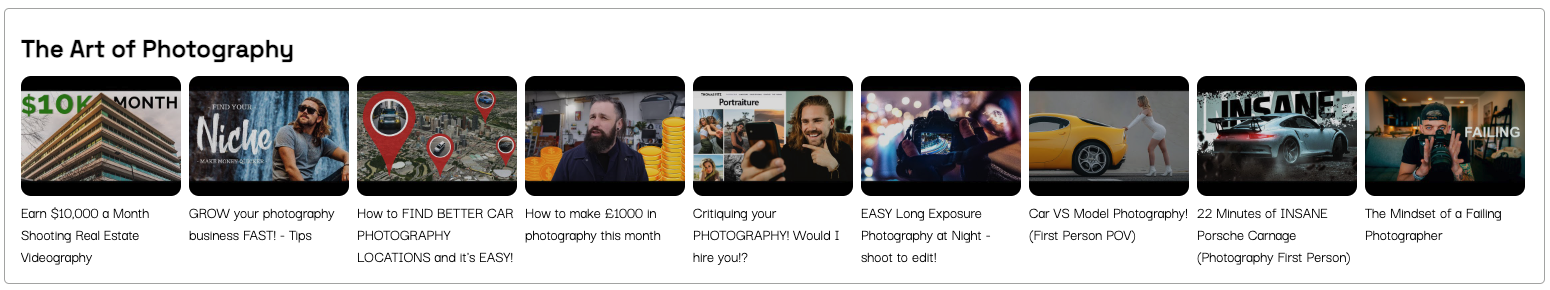}
    \includegraphics[width=0.9\textwidth]{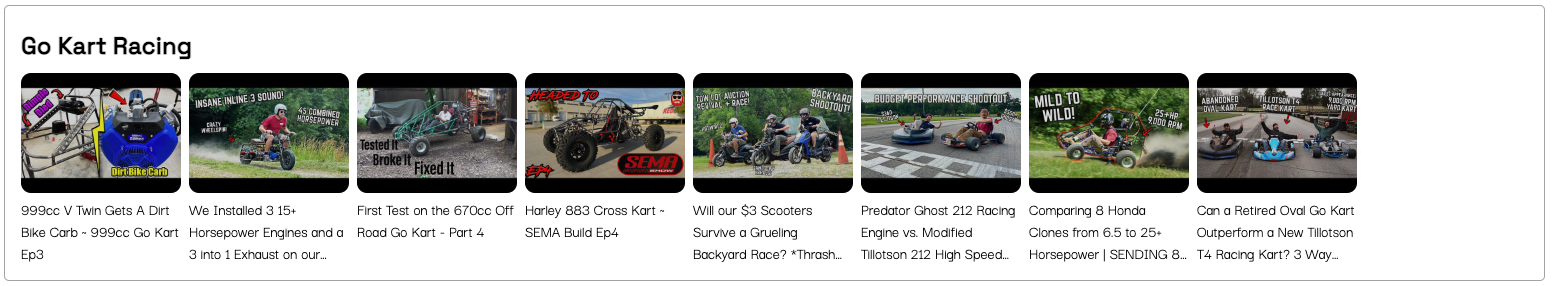}
    \caption{Examples of real extracted and named user journeys through our journey service.
    }
    \label{fig:example_qualitative}
    \vspace{-0.2in}
\end{figure}

\subsection{Key Result: Prompting LLMs can reason well through user interest journeys.}
As shown in Figure \ref{fig:example_qualitative},
our journey naming service can provide accurate, nuanced names describing the user interest journeys, like \emph{the art of photography}, or \emph{go kart racing}. These qualitative results illustrated are from a prompt-tuned LaMDA-137B model using expert-curated collections as explained in Section~\ref{subsec:data_alignment} (prompt tuning data), used in inference mode on top of real user journeys on an industrial recommendation platform, extracted via our ICPC method.

\subsection{RQ1: Which prompting technique performs  best for journey naming?}

\subsubsection*{Prompt tuning on small high quality data outperforms few-shot prompt engineering}



\begin{figure}[!ht]
    \centering
    \includegraphics[width=0.33\textwidth]{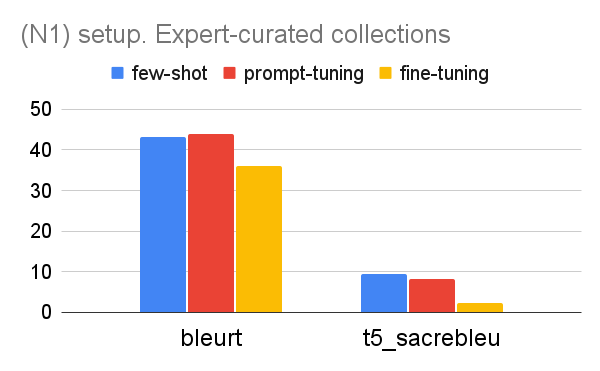}
\includegraphics[width=0.33\textwidth]{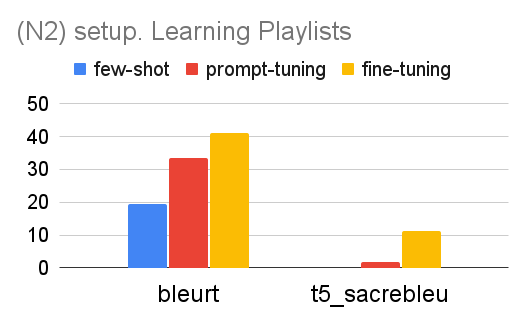}
    \includegraphics[width=0.33\textwidth]{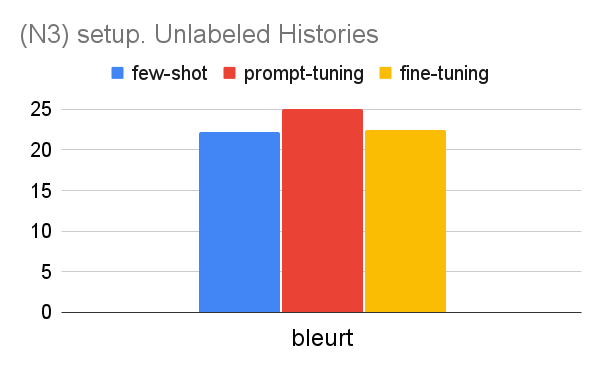}
    \caption{\textbf{Few shot vs. prompt tuning vs. fine-tuning}. (left) Expert collections. (middle) Learning playlists. (right) Unlabeled Histories.}
    \label{fig:larger_fs_vs_pt}  
\end{figure}
Figure \ref{fig:larger_fs_vs_pt}(middle) and (right) compares the different prompting techniques for LaMDA 137B in the larger experimental setups (N2) and (N3), respectively. We can see that for both setups, prompt-tuning performs significantly better compared to few-shot, as measured both by SacreBLEU and BLEURT scores. In Figure \ref{fig:larger_fs_vs_pt}(left) we also find similar results in the smaller but higher quality evaluation (N1) setup of the 100 expert-curated named collections: prompt-tuning outperforms few-shot prompt engineering both in terms of SacreBLEU and in BLEURT score. 
We also found that few-shot with 5 examples (bleurt 43.28) significantly outperforms zero-shot prompting (bleurt score 31.68). 


\subsubsection*{Fine-tuning outperforms prompt-tuning in-domain, but prompt-tuning has better generalization capability.}


Fine-tuning is the default strategy for aligning LLMs to new domains. We tested whether fine-tuning LaMDA 137B on 20,000 user-generated playlists (with input the items and output the corresponding user-given names) could outperform learning the prompt embedding based on the  (D3) smaller dataset of expertly curated named collections. The answer is: it depends. Based on results in Figure \ref{fig:larger_fs_vs_pt}, we find that fine-tuning outperforms prompt tuning for the (N2) evaluation setup where examples come from the same distribution as fine-tuning examples, aka Learning Playlists data. However, testing the generalization power of both prompting techniques in (N3) setup of unlabeled histories, prompt tuning outperforms fine-tuning. Also, from setup (N1) we find that  prompt-tuning with small in-domain examples outperforms fine-tuning with out-of-domain large scale data.



\subsection{RQ2: Which underlying model is better, and under which circumstances?}

\subsubsection*{Effect of model size}
\begin{figure}
    \centering
    \includegraphics[width=0.33\textwidth]{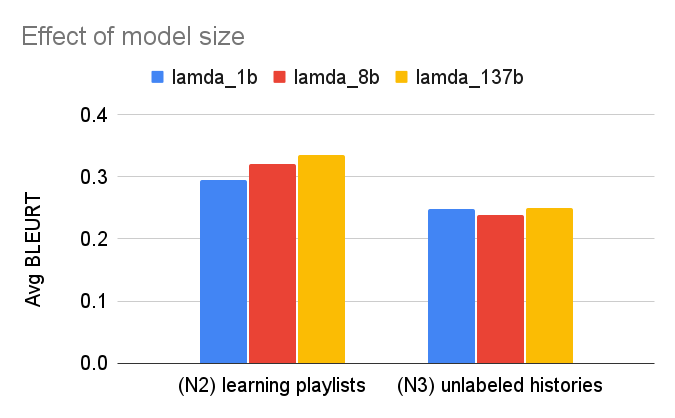}
    \includegraphics[width=0.33\textwidth]{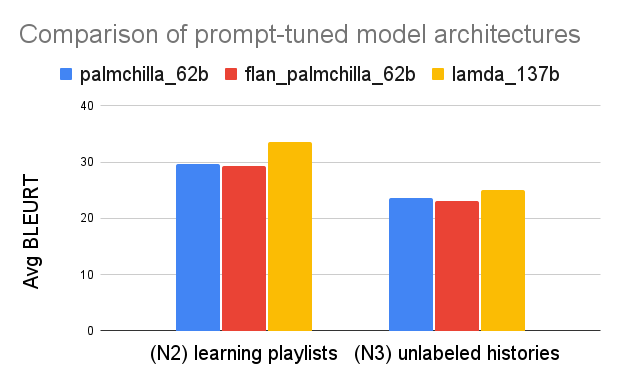}
    \includegraphics[width=0.33\textwidth]{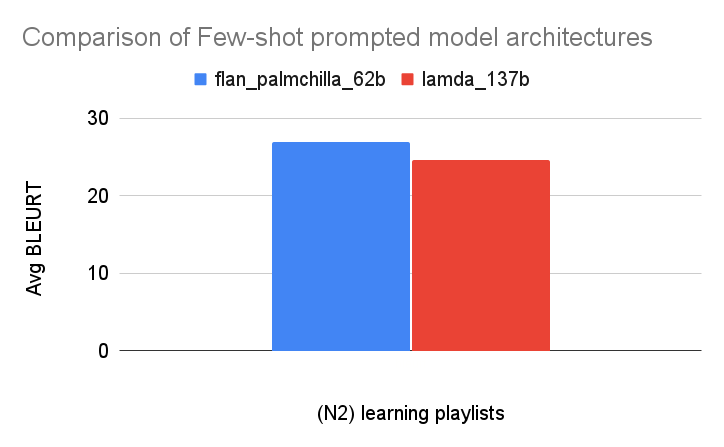}

    \caption{(left) \textbf{Effect of model size.} (middle, right) \textbf{Effect of model architectures.}}
    \label{fig:effect_model_size}
\vspace{-0.1in}
\end{figure}

Figure \ref{fig:effect_model_size}(left) compares how prompt tuning different sized LaMDA models affects the quality of generated journey names, considering the large-scale setups (N2) and (N3). One surprising observation is that while the performance on (N2) learning playlists suggests prompt tuning larger sized model leads to better naming quality, there is not a monotonic improvement on the (N3) unlabeled histories case. 
Overall, we explain these results by hypothesizing that when increasing the model size, on the one hand, the model has more capacity to infer good quality labels; on the other hand, the conditioning of model on the learned prompt embedding might become less prominent. Another hypothesis is the difference in the data among the two setups: (N3) compares output to input, while (N2) compares output to human labels. This trade-off among model size and prompt tuning needs further research investigation.

\subsubsection*{Effect of different model architectures}
In Figure \ref{fig:effect_model_size}(middle), we compare prompt-tuned models: PaLMChilla 62B, and its corresponding instruction tuned variant Flan-PaLMChilla 62B, with LaMDA 137B. We see that for both (N2) and (N3) setups, LaMDA 137B achieves the best BLEURT score. We also see that the performance of instruction-tuned PaLMchilla does not differ much from its non-instruction-tuned counterpart. 
Interestingly, when comparing few-shot prompted models FLAN PaLMChilla 62B with LaMDA 137B (Figure \ref{fig:effect_model_size}(right)), the order is reversed: FLAN PaLMChilla  outperforms LaMDA in (N2). We think that the instruction tuning of the former makes it better in following the few-shot prompting. Either way, both few-shot prompted models achieve lower BLEURT compared to the worst prompt-tuned model in the (N2) setup shown in Figure \ref{fig:effect_model_size} (middle).


\subsection*{RQ3: How construction of the prompt affects the quality of generated interest names?}

\subsubsection*{Which metadata are useful to be part of prompt?} 
\begin{figure}
    \centering
    \begin{minipage}{0.33\linewidth}
    \includegraphics[width=\textwidth]{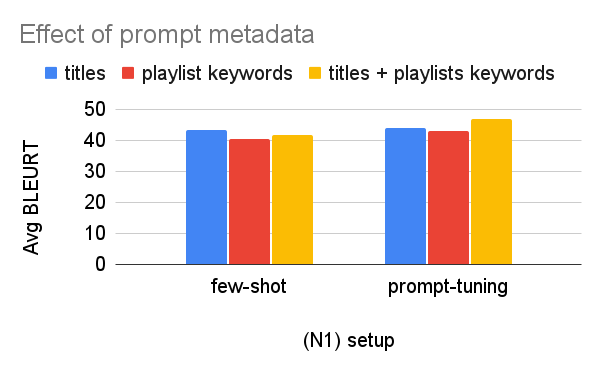}
    \caption{Effect of  prompt metadata.}
    \label{fig:effect_prompt_metadata}  
    \end{minipage}
    \begin{minipage}{0.33\linewidth}
    \includegraphics[width=\textwidth]{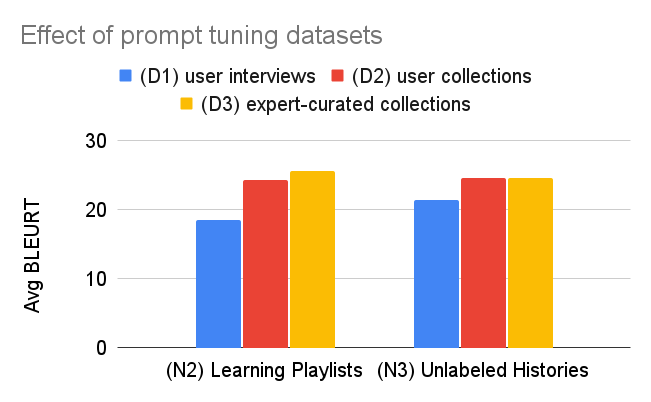}
    \caption{Effect of prompt-tuning datasets.}
    \label{fig:effect_prompt_tuning_data}  
    \end{minipage}
 \begin{minipage}{0.33\linewidth}
    \includegraphics[width=\textwidth]{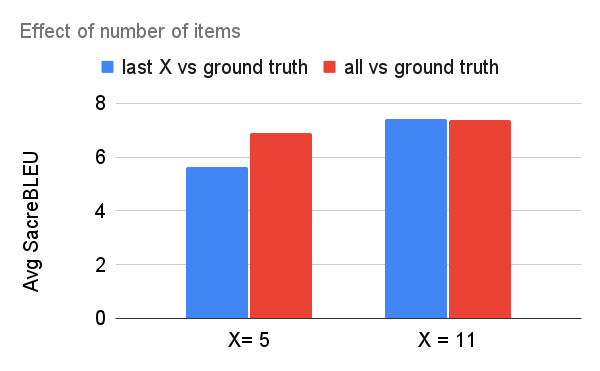}
    \caption{Effect of number of journey items. 
    }
    \label{fig:journey_length}
    \end{minipage}
    \vspace{-0.1in}
\end{figure}

Figure \ref{fig:effect_prompt_metadata} shows for (N1) setup, how different ways of representing the items: (1) using item titles/names, or (2) using keywords representing the playlist, or (3) with both item names and playlist keywords (as discussed in Table \ref{tab:format_examples} for structured prompts) interact with different prompting techniques. Overall, we find that the titles of the items is the most informative feature, with some additive value provided when including playlist keywords. 
Therefore, unless specified otherwise, we infer journey name given only titles metadata.



\subsubsection*{Effect of prompt tuning data on success of learned prompt embeddings.}

As expected, the quality of the data used to learn the prompt embedding plays a key role in how successful prompt tuning LLMs can be. In Figure \ref{fig:effect_prompt_tuning_data}, we compare how learning the prompt on top of three different small quality datasets generalizes in the task of inferring journey names for the larger (N2) and (N3) experimental tasks. The prompt tuning data are the (D1) user interviews, (D2) user collections, and (D3) expert-curated collections, as described in Section \ref{subsec:alignllms}. 
In both (N2) and (N3) setups, we see a big difference in performance based on the data we use to learn the prompt, with prompt embeddings learned from expertly curated names of collections providing the highest generalization capability. We argue that this is the case because the expertly named collections provides higher quality data to learn the prompt compared to user provided names, which tend to be more noisy, especially in the context of an interview where recipients might provide data of mixed quality. These results also point to the potential of mixed methods, i.e., mixing expert-provided labels, with a subset of higher quality user-provided labels, for enabling LLMs to accurately describe user interests.  

\subsubsection*{Effect of journey length.} We also conduct experiments to understand the number of items in the journey that should be included in order to generate a good name. Figure \ref{fig:journey_length} shows that if we generate names only given the last five items in the journey, the SacreBLEU score between ground truth and generated name is significantly lower compared to if we were to use all the items in the journey\footnote{This comparison is done limiting to journeys that have more than five items, in the (N2) setup.}. However, when comparing journey names as inferred by the last eleven items only (now in histories of more than eleven items), there is little drop in performance compared to using all items in journeys. This can be a guiding point to reduce the context length in generating journey names. 


\subsection{RQ4: How safe and interesting generated user journey names are?}

As pioneered in \cite{thoppilan2022lamda}, besides the accuracy of responses generated by LLMs, other evaluation dimensions should be considered as well. These include how safe/responsible the responses are; how interesting they are; whether they are sensible; and whether they are specific enough, with many of these dimensions competing one against the another (e.g., the safer the response, the less interesting it might be). 
We utilize the already trained LaMDA scorer \cite{thoppilan2022lamda} (without the prompting done to align to the journey interest domain), as the imputation model for the aforementioned dimensions. . 
Figure \ref{fig:safety} plots the imputed scores for the generated, by the prompt-tuned LaMDA-137B model, names, in the three setups (N1), (N2) and (N3). Our key takeaway is that the degree to which the generated journey names are safe/ interesting/ and so on, is largely affected by the underlying input given in inference to produce the names. This is evident by the fact that the more in-the-wild unlabeled histories have names which have lower safety score compared to names produced by the same model for the user-curated playlists, or the expert-collections. 



\subsection{RQ5: Do we need  journey extraction, or can we rely on LLMs on both journey extraction \& naming?}

\begin{figure*}
    \begin{minipage}{0.33\linewidth}
    \includegraphics[width=\textwidth]{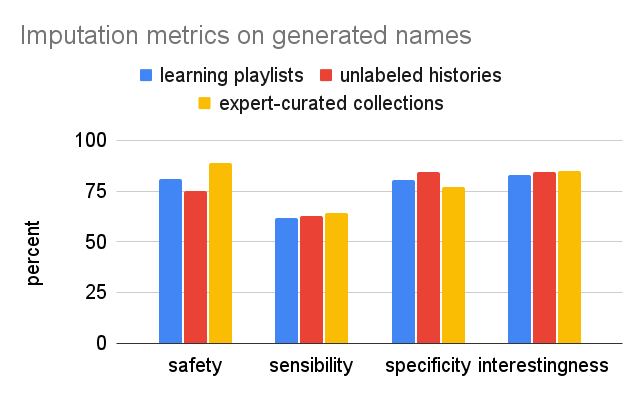}
    \caption{Imputed aspects for names. 
    }
    \label{fig:safety}
    \end{minipage}
  \begin{minipage}{0.33\textwidth}
    \centering
    \includegraphics[width=\textwidth]{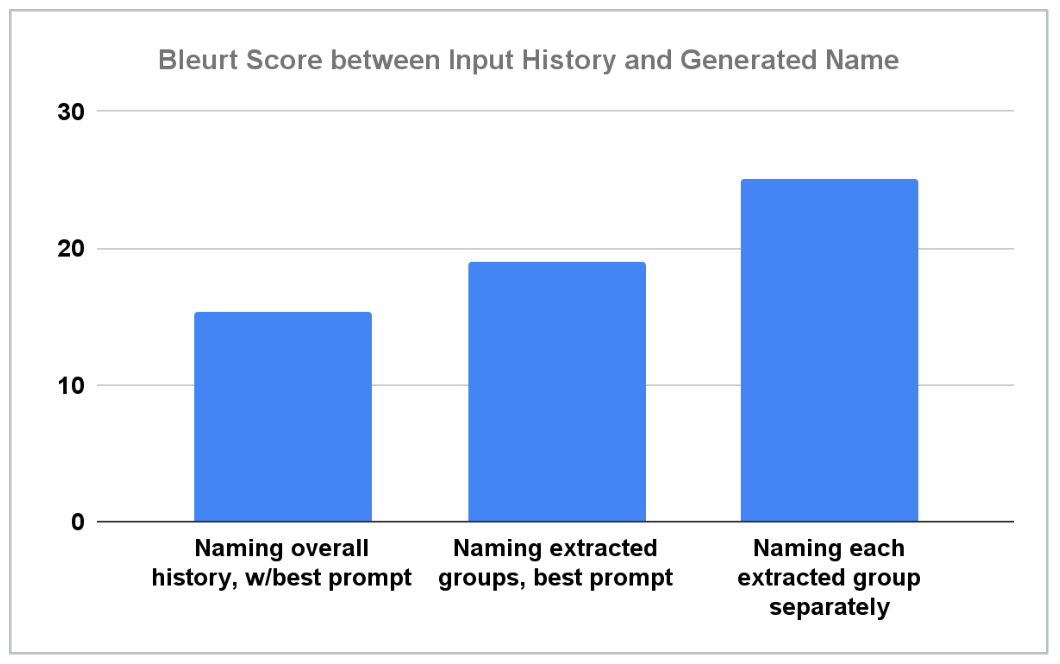}
    \captionof{figure}{Value of extracting journeys, separate than naming.}
    \label{fig:extraction_value}
  \end{minipage}
  \begin{minipage}{0.33\textwidth}
    \centering
    \includegraphics[width=\textwidth]{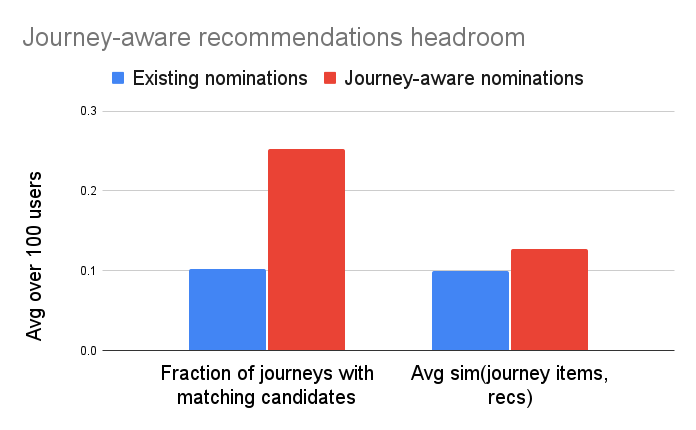}
    \captionof{figure}{Headroom of journey-aware recommendations.}
    \label{fig:recommendations_headroom}
  \end{minipage}
  \vspace{-0.1in}
\end{figure*}

  
A natural question that might arise to the readers is that given LLMs evidently can be used to generate nuanced interesting names of the interest journeys users are pursuing, why not also task them with the job of both extracting and naming simultaneously? To answer this question, we use the evaluation setup (N3). 
The approaches we compare are: (1) Providing the whole user history to the LLM and asking the model to name the interest journeys the user is pursuing. (2) Extracting journeys using our ICPC, and concatenating the journey groups (compared to the mixed history in (1)) at once in a single inference to the LLM and asking it to name the journey groups. (3) Naming each extracted group separately, with a different LLM call per journey. We compute the BLEURT score between the whole input history and the concatenated journey names as the final evaluation score and compare the three approaches. It's evident from Figure \ref{fig:extraction_value} that the LLM can generate significantly better journey names when it's only given the journey-specific history, thus further validating the need for our journey extraction component in the journey service.

\subsection{RQ6: Can the extracted and named journeys enable an improved recommendation experience?}
Lastly, we provide some early analysis on the degree to which a user's journeys can be better assisted when the recommender is aware of the underlying journeys. We compare the  journeys extracted from real user histories over the period of a month (same setup as N3) with the recommendations given to the user the day after the month on the industrial platform. For each journey of a user, we compute the similarity of the average salient terms embedding of the journey and that of all the recommended items\footnote{For computation purposes, we reduce comparison of each journey with each recommended item, to comparing with the entire recommendation set}. 
When the similarity is larger than a threshold (set to 0.2), we track the journey as being served. We can then compare this with a new recommendation process using our journey service and search: 1) we extract journeys from the one-month user history and name the extracted journeys using prompt-tuned LaMDA 137B on (D3); 2) we generate recommendations for each journey through searching the platform using the generated journey name. 
We then compute the similarities between these new journey-aware recommendations and journeys for comparison.  
We can see in Figure \ref{fig:recommendations_headroom} that on average over a hundred sampled users, 
around 1 out of 10 journeys is served (0.102) under the current recommendation system. This number is significantly improved when the recommender is journey-aware (0.253). We also report the average similarity score between journeys and candidates (without the thresholding). We can see that overall, the provided recommedation candidates have higher  similarity in the case of journey aware nominations (0.127 vs 0.099). Of course, to fully answer this question, ideally we need novel recommendation approaches mapping the journeys to recommendations. We also need A/B user facing experiments to measure the value added to the user. Both will be part of our future work. Nevertheless, these early results already point to headroom of improving recommendations via making them aware of user interest journeys.




%% file: formulation.tex





\textbf{Intent, Tasks, User Needs} 
We compare journeys with related work on  understanding and modeling {user needs}, {intent}, and {tasks}. Intents has long been considered in the search domain \cite{zhang2011user, radlinski2010inferring}. 
Recent works on intent-aware recommendation  \cite{bhattacharya2017intent, mehrotra2019jointly} shows modeling intent improves user satisfaction. 
While intent captures the high-level context of the user’s visit, it is much coarser than journeys. For example, the intent of "I want to do something inspiring" covers “becoming an entrepreneur” and “learning the ukulele” journeys. 
Another related work is modeling task behavior, which goes beyond the current single query, in search engines to improve search ranking (e.g., \cite{white2013enhancing, lucchese2011identifying}). However, search tasks are typically based on in-session interactions \cite{lucchese2011identifying} or across  only a few sessions \cite{kotov2011modeling}, while users are found to pursue interest journeys for a month, or more. In addition, interest journeys also cover persisting interest such as entertainment, which is less task-oriented. 
User needs in Web search were categorized in \citep{rose2004understanding, broder2002taxonomy, yin2010building} as transactional, navigational, or informational. A similar categorization on \emph{why} users utilize recommendation systems has not been defined. Our findings suggest a new dimension of user needs: the dimension of using recommenders alongside real-life interests and aspirations, i.e., learning new skills, entertainment, and sense of belonging. This confirms our insight that the recommendation process should be transformed to satisfy this additional need. Broadly speaking, the goals of journey-aware recommendation align with those of personal assistants \cite{guha2015user, sun2016contextual, song2016query, safavi2020toward} and context-aware recommenders \cite{baltrunas2011matrix}. Our journey extraction technology can be utilized as context within a proactive personal assistant to tailor recommendations.

\textbf{User Interests Modeling}  Another close line of work is user modeling in recommender systems, and particularly modeling changing user interests via sequential models (e.g., RNNs) on user actions \cite{wu2017recurrent, chen2019top}. 
Although powerful, sequence models such as RNNs tend to focus only on the last interactions, forgetting long-term user interests. 
To mitigate the recency bias, a recent thread has focused on incorporating users’ long-term history data into recommendation models \cite{pancha2022pinnerformer, pi2019practice, zhou2019deep} by powering models with attention-based mechanisms \cite{kang2018self, sun2019bert4rec}. The produced user representation however is hard to examine or interpret. In this work, we take a different approach: extracting longer-term valued journeys, separately from the recommendation task. This allows for modularity of involved components and for interpretability  \cite{radlinski2022natural}, to enable the system explain recommendations via the lens of journeys \cite{tintarev2007explanations}.
 Our journey extraction service allows to easily inspect and  control the granularity of extracted journeys, while naming each journey via LLMs makes our approach interpretable.  Furthermore, by introducing the journey entity, we can eventually enable users to interact with their extracted journeys; e.g., control whether they want more or less recommendations related to that journey. 
Also, our approach of extracting journeys via clustering, while paying attention to content understanding is related to user topic modeling \cite{zaheer2017latent, bahrainian2019predicting, 37745} and clustering content by user needs \cite{kong2016improving, 49265}. Also, works on uncovering coherent trails \cite{chi2001using, shahaf2010connecting, shahaf2012trains, 49265} inspired us greatly.

\textbf{Interpretable/Transparent User Models} Our work is related to many past works advocating for the need of \emph{explainable recommendation} \cite{tintarev2007explanations}, as well as \emph{user control} \cite{jannach2017user}. Our contribution is the first-ever study showing the power of prompted LLMs so that they can reason through user interests; as well as shedding light to the angle of persistent real-life user journeys that users expect recommenders to help them with besides the shorter-term tasks. Perhaps the most close work to ours is 
\cite{radlinski2022natural} which laid out the vision for
scrutable Natural Language (NL)-based recommendation. 
Our work however has several differences: Rather than a complete summary of user interests, we aim at short sentences describing different user interest journeys. Also, while \cite{radlinski2022natural} focuses on scrutable profiles, with the aim of recommendation, we focus on investigating the degree to which our personalized clustering approach coupled with prompting LLMs can provide deeper user understanding. Perhaps the most important difference is that, to our knowledge, our work is the first-ever experimental study of aligning LLMs to reason through user journeys. Furthermore, our work differs significantly from \cite{balog2019transparent} where the authors showed how a restricted representation of users as weighted pairs of tag interaction can be verbalized as NL statements using templates. Our models rely on free-form natural language, rather than templates to output interest names.

\textbf{Large Language Models (LLMs)} Last but not least, our work is the first to demonstrate the application of LLMs in the user interest journey domain, for deeper user understanding in recommendation systems. While \cite{radlinski2022natural} provided some key directions along the usage of LLMs in the recommendation space as mentioned above, to our knowledge this is the first study to address issues like which data to use to fine-tune and prompt-tune LLMs to align them with this domain, how different prompting techniques perform, among others. We argue that there is large untapped potential in LLMs for the domain of recommender systems. We know that recent large pre-trained language models have been shown to have impressive generative fluency \cite{thoppilan2022lamda, chowdhery2022palm} as well as effective few-shot learning capabilities \cite{brown2020language} in the domain of natural language generation and understanding, and have passed several benchmarks of various tasks. However, so far, the key application to the recommendation system domain has been the P5 paper \cite{geng2022recommendation} which effectively maps recommendation as a language processing task, framing a variety of recommendation tasks into an end-to-end transformers framework. 
Our work is orthogonal; it leverages the power of LLMs as a way to reason through user interests, so to provide the interpretability a human would offer if they were to be asked for a recommendation.

%% file: conclusions.tex
Our work aims to bridge the gap in LLMs for understanding user interest journeys and improving user experience on recommendation platforms. We provide the first-ever demonstration of LLMs to reason through user interests and describe them similarly to how people would. We uncover important aspects to enable such capabilities, such as data, prompting techniques, and evaluation methodology. 
We show that prompt-tuning LLMs based on small but quality data of expertly curated and named collections has the most generalization capability. We also find first extracting the interest journeys through a personalized clustering procedure is critical. 
We validate our findings on a large-scale industrial recommender platform, and together suggest the promise of utilizing our proposed journey service as a way to gain deeper user understanding and building novel user experiences that offer more user control. 

%% file: sample-xelatex.bbl

%% file: sample-xelatex.bbl

\begin{thebibliography}{60}


\ifx \showCODEN    \undefined \def \showCODEN     #1{\unskip}     \fi
\ifx \showDOI      \undefined \def \showDOI       #1{#1}\fi
\ifx \showISBNx    \undefined \def \showISBNx     #1{\unskip}     \fi
\ifx \showISBNxiii \undefined \def \showISBNxiii  #1{\unskip}     \fi
\ifx \showISSN     \undefined \def \showISSN      #1{\unskip}     \fi
\ifx \showLCCN     \undefined \def \showLCCN      #1{\unskip}     \fi
\ifx \shownote     \undefined \def \shownote      #1{#1}          \fi
\ifx \showarticletitle \undefined \def \showarticletitle #1{#1}   \fi
\ifx \showURL      \undefined \def \showURL       {\relax}        \fi
\providecommand\bibfield[2]{#2}
\providecommand\bibinfo[2]{#2}
\providecommand\natexlab[1]{#1}
\providecommand\showeprint[2][]{arXiv:#2}

\bibitem[beu({[n.\,d.]})]%
        {beutel2018latent}
 \bibinfo{year}{[n.\,d.]}\natexlab{}.
\newblock


\bibitem[Bahrainian et~al\mbox{.}(2019)]%
        {bahrainian2019predicting}
\bibfield{author}{\bibinfo{person}{Seyed~Ali Bahrainian},
  \bibinfo{person}{Fattane Zarrinkalam}, \bibinfo{person}{Ida Mele}, {and}
  \bibinfo{person}{Fabio Crestani}.} \bibinfo{year}{2019}\natexlab{}.
\newblock \showarticletitle{Predicting the topic of your next query for
  just-in-time ir}. In \bibinfo{booktitle}{\emph{European Conference on
  Information Retrieval}}. Springer, \bibinfo{pages}{261--275}.
\newblock


\bibitem[Balog et~al\mbox{.}(2019)]%
        {balog2019transparent}
\bibfield{author}{\bibinfo{person}{Krisztian Balog}, \bibinfo{person}{Filip
  Radlinski}, {and} \bibinfo{person}{Shushan Arakelyan}.}
  \bibinfo{year}{2019}\natexlab{}.
\newblock \showarticletitle{Transparent, scrutable and explainable user models
  for personalized recommendation}. In \bibinfo{booktitle}{\emph{Proceedings of
  the 42nd international acm sigir conference on research and development in
  information retrieval}}. \bibinfo{pages}{265--274}.
\newblock


\bibitem[Baltrunas et~al\mbox{.}(2011)]%
        {baltrunas2011matrix}
\bibfield{author}{\bibinfo{person}{Linas Baltrunas}, \bibinfo{person}{Bernd
  Ludwig}, {and} \bibinfo{person}{Francesco Ricci}.}
  \bibinfo{year}{2011}\natexlab{}.
\newblock \showarticletitle{Matrix factorization techniques for context aware
  recommendation}. In \bibinfo{booktitle}{\emph{Proceedings of the fifth ACM
  conference on Recommender systems}}. \bibinfo{pages}{301--304}.
\newblock


\bibitem[Bhattacharya et~al\mbox{.}(2017)]%
        {bhattacharya2017intent}
\bibfield{author}{\bibinfo{person}{Biswarup Bhattacharya},
  \bibinfo{person}{Iftikhar Burhanuddin}, \bibinfo{person}{Abhilasha Sancheti},
  {and} \bibinfo{person}{Kushal Satya}.} \bibinfo{year}{2017}\natexlab{}.
\newblock \showarticletitle{Intent-aware contextual recommendation system}. In
  \bibinfo{booktitle}{\emph{2017 IEEE International Conference on Data Mining
  Workshops (ICDMW)}}. IEEE, \bibinfo{pages}{1--8}.
\newblock


\bibitem[Broder(2002)]%
        {broder2002taxonomy}
\bibfield{author}{\bibinfo{person}{Andrei Broder}.}
  \bibinfo{year}{2002}\natexlab{}.
\newblock \showarticletitle{A taxonomy of web search}. In
  \bibinfo{booktitle}{\emph{ACM Sigir forum}}, Vol.~\bibinfo{volume}{36}. ACM
  New York, NY, USA, \bibinfo{pages}{3--10}.
\newblock


\bibitem[Cer et~al\mbox{.}(2018)]%
        {cer2018universal}
\bibfield{author}{\bibinfo{person}{Daniel Cer}, \bibinfo{person}{Yinfei Yang},
  \bibinfo{person}{Sheng-yi Kong}, \bibinfo{person}{Nan Hua},
  \bibinfo{person}{Nicole Limtiaco}, \bibinfo{person}{Rhomni~St John},
  \bibinfo{person}{Noah Constant}, \bibinfo{person}{Mario Guajardo-Cespedes},
  \bibinfo{person}{Steve Yuan}, \bibinfo{person}{Chris Tar}, {et~al\mbox{.}}}
  \bibinfo{year}{2018}\natexlab{}.
\newblock \showarticletitle{Universal sentence encoder}.
\newblock \bibinfo{journal}{\emph{arXiv preprint arXiv:1803.11175}}
  (\bibinfo{year}{2018}).
\newblock


\bibitem[Chen et~al\mbox{.}(2019)]%
        {chen2019top}
\bibfield{author}{\bibinfo{person}{Minmin Chen}, \bibinfo{person}{Alex Beutel},
  \bibinfo{person}{Paul Covington}, \bibinfo{person}{Sagar Jain},
  \bibinfo{person}{Francois Belletti}, {and} \bibinfo{person}{Ed~H Chi}.}
  \bibinfo{year}{2019}\natexlab{}.
\newblock \showarticletitle{Top-k off-policy correction for a REINFORCE
  recommender system}. In \bibinfo{booktitle}{\emph{Proceedings of the Twelfth
  ACM International Conference on Web Search and Data Mining}}.
  \bibinfo{pages}{456--464}.
\newblock


\bibitem[Chi et~al\mbox{.}(2001)]%
        {chi2001using}
\bibfield{author}{\bibinfo{person}{Ed~H Chi}, \bibinfo{person}{Peter Pirolli},
  \bibinfo{person}{Kim Chen}, {and} \bibinfo{person}{James Pitkow}.}
  \bibinfo{year}{2001}\natexlab{}.
\newblock \showarticletitle{Using information scent to model user information
  needs and actions and the Web}. In \bibinfo{booktitle}{\emph{Proceedings of
  the SIGCHI conference on Human factors in computing systems}}.
  \bibinfo{pages}{490--497}.
\newblock


\bibitem[Chowdhery(2022)]%
        {chowdhery2022palm}
\bibfield{author}{\bibinfo{person}{Aakanksha et~al. Chowdhery}.}
  \bibinfo{year}{2022}\natexlab{}.
\newblock \showarticletitle{Palm: Scaling language modeling with pathways}.
\newblock \bibinfo{journal}{\emph{arXiv preprint arXiv:2204.02311}}
  (\bibinfo{year}{2022}).
\newblock


\bibitem[Covington et~al\mbox{.}(2016)]%
        {covington2016deep}
\bibfield{author}{\bibinfo{person}{Paul Covington}, \bibinfo{person}{Jay
  Adams}, {and} \bibinfo{person}{Emre Sargin}.}
  \bibinfo{year}{2016}\natexlab{}.
\newblock \showarticletitle{Deep neural networks for youtube recommendations}.
  In \bibinfo{booktitle}{\emph{Proceedings of the 10th ACM conference on
  recommender systems}}. \bibinfo{pages}{191--198}.
\newblock


\bibitem[Ekstrand and Willemsen(2016)]%
        {ekstrand2016behaviorism}
\bibfield{author}{\bibinfo{person}{Michael~D Ekstrand} {and}
  \bibinfo{person}{Martijn~C Willemsen}.} \bibinfo{year}{2016}\natexlab{}.
\newblock \showarticletitle{Behaviorism is not enough: better recommendations
  through listening to users}. In \bibinfo{booktitle}{\emph{Proceedings of the
  10th ACM conference on recommender systems}}. \bibinfo{pages}{221--224}.
\newblock


\bibitem[et~al.(2022a)]%
        {chung2022scaling}
\bibfield{author}{\bibinfo{person}{Hyung Won~Chung et al.}}
  \bibinfo{year}{2022}\natexlab{a}.
\newblock \bibinfo{title}{Scaling Instruction-Finetuned Language Models}.
\newblock
\newblock
\showeprint[arxiv]{2210.11416}~[cs.LG]


\bibitem[et~al.(2022b)]%
        {hoffmann2022training}
\bibfield{author}{\bibinfo{person}{Jordan~Hoffmann et al.}}
  \bibinfo{year}{2022}\natexlab{b}.
\newblock \bibinfo{title}{Training Compute-Optimal Large Language Models}.
\newblock
\newblock
\showeprint[arxiv]{2203.15556}~[cs.CL]


\bibitem[et~al.(2022c)]%
        {barham2022pathways}
\bibfield{author}{\bibinfo{person}{Paul~Barham et al.}}
  \bibinfo{year}{2022}\natexlab{c}.
\newblock \bibinfo{title}{Pathways: Asynchronous Distributed Dataflow for ML}.
\newblock
\newblock
\showeprint[arxiv]{2203.12533}~[cs.DC]


\bibitem[et~al.(2020)]%
        {brown2020language}
\bibfield{author}{\bibinfo{person}{Tom B.~Brown et al.}}
  \bibinfo{year}{2020}\natexlab{}.
\newblock \bibinfo{title}{Language Models are Few-Shot Learners}.
\newblock
\newblock
\showeprint[arxiv]{2005.14165}~[cs.CL]


\bibitem[Gamon et~al\mbox{.}(2016)]%
        {gamon2016identifying}
\bibfield{author}{\bibinfo{person}{Michael Gamon}, \bibinfo{person}{Patrick
  Pantel}, \bibinfo{person}{Xinying Song}, \bibinfo{person}{Tae Yano}, {and}
  \bibinfo{person}{Johnson~Tan Apacible}.} \bibinfo{year}{2016}\natexlab{}.
\newblock \bibinfo{title}{Identifying salient items in documents}.
\newblock
\newblock
\newblock
\shownote{US Patent 9,251,473}.


\bibitem[Geng et~al\mbox{.}(2022)]%
        {geng2022recommendation}
\bibfield{author}{\bibinfo{person}{Shijie Geng}, \bibinfo{person}{Shuchang
  Liu}, \bibinfo{person}{Zuohui Fu}, \bibinfo{person}{Yingqiang Ge}, {and}
  \bibinfo{person}{Yongfeng Zhang}.} \bibinfo{year}{2022}\natexlab{}.
\newblock \showarticletitle{Recommendation as language processing (rlp): A
  unified pretrain, personalized prompt \& predict paradigm (p5)}. In
  \bibinfo{booktitle}{\emph{Proceedings of the 16th ACM Conference on
  Recommender Systems}}. \bibinfo{pages}{299--315}.
\newblock


\bibitem[Guha et~al\mbox{.}(2015)]%
        {guha2015user}
\bibfield{author}{\bibinfo{person}{Ramanathan Guha}, \bibinfo{person}{Vineet
  Gupta}, \bibinfo{person}{Vivek Raghunathan}, {and}
  \bibinfo{person}{Ramakrishnan Srikant}.} \bibinfo{year}{2015}\natexlab{}.
\newblock \showarticletitle{User modeling for a personal assistant}. In
  \bibinfo{booktitle}{\emph{Proceedings of the Eighth ACM International
  Conference on Web Search and Data Mining}}. \bibinfo{pages}{275--284}.
\newblock


\bibitem[He et~al\mbox{.}(2017)]%
        {he2017neural}
\bibfield{author}{\bibinfo{person}{Xiangnan He}, \bibinfo{person}{Lizi Liao},
  \bibinfo{person}{Hanwang Zhang}, \bibinfo{person}{Liqiang Nie},
  \bibinfo{person}{Xia Hu}, {and} \bibinfo{person}{Tat-Seng Chua}.}
  \bibinfo{year}{2017}\natexlab{}.
\newblock \bibinfo{title}{Neural Collaborative Filtering}.
\newblock
\newblock
\showeprint[arxiv]{1708.05031}~[cs.IR]


\bibitem[Jannach et~al\mbox{.}(2017)]%
        {jannach2017user}
\bibfield{author}{\bibinfo{person}{Dietmar Jannach}, \bibinfo{person}{Sidra
  Naveed}, {and} \bibinfo{person}{Michael Jugovac}.}
  \bibinfo{year}{2017}\natexlab{}.
\newblock \showarticletitle{User control in recommender systems: Overview and
  interaction challenges}. In \bibinfo{booktitle}{\emph{E-Commerce and Web
  Technologies: 17th International Conference, EC-Web 2016, Porto, Portugal,
  September 5-8, 2016}}. Springer, \bibinfo{pages}{21--33}.
\newblock


\bibitem[Kang and McAuley(2018)]%
        {kang2018self}
\bibfield{author}{\bibinfo{person}{Wang-Cheng Kang} {and}
  \bibinfo{person}{Julian McAuley}.} \bibinfo{year}{2018}\natexlab{}.
\newblock \showarticletitle{Self-attentive sequential recommendation}. In
  \bibinfo{booktitle}{\emph{2018 IEEE international conference on data mining
  (ICDM)}}. IEEE, \bibinfo{pages}{197--206}.
\newblock


\bibitem[Kong et~al\mbox{.}(2016)]%
        {kong2016improving}
\bibfield{author}{\bibinfo{person}{Jing Kong}, \bibinfo{person}{Alex Scott},
  {and} \bibinfo{person}{Georg~M Goerg}.} \bibinfo{year}{2016}\natexlab{}.
\newblock \showarticletitle{Improving topic clustering on search queries with
  word co-occurrence and bipartite graph co-clustering}.
\newblock  (\bibinfo{year}{2016}).
\newblock


\bibitem[Kong et~al\mbox{.}(2020)]%
        {49265}
\bibfield{author}{\bibinfo{person}{Weize Kong}, \bibinfo{person}{Mike
  Bendersky}, \bibinfo{person}{Marc Najork}, \bibinfo{person}{Brandon Vargo},
  {and} \bibinfo{person}{Mike Colagrosso}.} \bibinfo{year}{2020}\natexlab{}.
\newblock \showarticletitle{Learning to Cluster Documents into Workspaces Using
  Large Scale Activity Logs}. In \bibinfo{booktitle}{\emph{Proceedings of the
  26th ACM SIGKDD Conference on Knowledge Discovery and Data Mining (KDD
  ’20)}}. \bibinfo{pages}{2416–2424}.
\newblock


\bibitem[Koren et~al\mbox{.}(2009)]%
        {koren2009matrix}
\bibfield{author}{\bibinfo{person}{Yehuda Koren}, \bibinfo{person}{Robert
  Bell}, {and} \bibinfo{person}{Chris Volinsky}.}
  \bibinfo{year}{2009}\natexlab{}.
\newblock \showarticletitle{Matrix factorization techniques for recommender
  systems}.
\newblock \bibinfo{journal}{\emph{Computer}} \bibinfo{volume}{42},
  \bibinfo{number}{8} (\bibinfo{year}{2009}), \bibinfo{pages}{30--37}.
\newblock


\bibitem[Kotov et~al\mbox{.}(2011)]%
        {kotov2011modeling}
\bibfield{author}{\bibinfo{person}{Alexander Kotov}, \bibinfo{person}{Paul~N
  Bennett}, \bibinfo{person}{Ryen~W White}, \bibinfo{person}{Susan~T Dumais},
  {and} \bibinfo{person}{Jaime Teevan}.} \bibinfo{year}{2011}\natexlab{}.
\newblock \showarticletitle{Modeling and analysis of cross-session search
  tasks}. In \bibinfo{booktitle}{\emph{Proceedings of the 34th international
  ACM SIGIR conference on Research and development in Information Retrieval}}.
  \bibinfo{pages}{5--14}.
\newblock


\bibitem[Lester et~al\mbox{.}(2021)]%
        {lester2021power}
\bibfield{author}{\bibinfo{person}{Brian Lester}, \bibinfo{person}{Rami
  Al-Rfou}, {and} \bibinfo{person}{Noah Constant}.}
  \bibinfo{year}{2021}\natexlab{}.
\newblock \bibinfo{title}{The Power of Scale for Parameter-Efficient Prompt
  Tuning}.
\newblock
\newblock
\showeprint[arxiv]{2104.08691}~[cs.CL]


\bibitem[Li et~al\mbox{.}(2011)]%
        {37686}
\bibfield{author}{\bibinfo{person}{Zhen Li}, \bibinfo{person}{Huazhong Ning},
  \bibinfo{person}{Liangliang Cao}, \bibinfo{person}{Tong Zhan},
  \bibinfo{person}{Yihong Gong}, {and} \bibinfo{person}{Thomas~S. Huang}.}
  \bibinfo{year}{2011}\natexlab{}.
\newblock \showarticletitle{Learning to Search Efficiently in High Dimensions}.
  In \bibinfo{booktitle}{\emph{Neural Information Processing Systems}}.
\newblock


\bibitem[Liang(2019)]%
        {liang2019recommender}
\bibfield{author}{\bibinfo{person}{Yu Liang}.} \bibinfo{year}{2019}\natexlab{}.
\newblock \showarticletitle{Recommender system for developing new preferences
  and goals}. In \bibinfo{booktitle}{\emph{Proceedings of the 13th ACM
  Conference on Recommender Systems}}. \bibinfo{pages}{611--615}.
\newblock


\bibitem[Liang et~al\mbox{.}(2023)]%
        {liang2023enabling}
\bibfield{author}{\bibinfo{person}{Yu Liang}, \bibinfo{person}{Aditya Ponnada},
  \bibinfo{person}{Paul Lamere}, {and} \bibinfo{person}{Nediyana Daskalova}.}
  \bibinfo{year}{2023}\natexlab{}.
\newblock \showarticletitle{Enabling Goal-Focused Exploration of Podcasts in
  Interactive Recommender Systems}. In \bibinfo{booktitle}{\emph{Proceedings of
  the 28th International Conference on Intelligent User Interfaces}}.
  \bibinfo{pages}{142--155}.
\newblock


\bibitem[Lucchese et~al\mbox{.}(2011)]%
        {lucchese2011identifying}
\bibfield{author}{\bibinfo{person}{Claudio Lucchese},
  \bibinfo{person}{Salvatore Orlando}, \bibinfo{person}{Raffaele Perego},
  \bibinfo{person}{Fabrizio Silvestri}, {and} \bibinfo{person}{Gabriele
  Tolomei}.} \bibinfo{year}{2011}\natexlab{}.
\newblock \showarticletitle{Identifying task-based sessions in search engine
  query logs}. In \bibinfo{booktitle}{\emph{Proceedings of the fourth ACM
  international conference on Web search and data mining}}.
  \bibinfo{pages}{277--286}.
\newblock


\bibitem[McInnes et~al\mbox{.}(2020)]%
        {mcinnes2020umap}
\bibfield{author}{\bibinfo{person}{Leland McInnes}, \bibinfo{person}{John
  Healy}, {and} \bibinfo{person}{James Melville}.}
  \bibinfo{year}{2020}\natexlab{}.
\newblock \bibinfo{title}{UMAP: Uniform Manifold Approximation and Projection
  for Dimension Reduction}.
\newblock
\newblock
\showeprint[arxiv]{1802.03426}~[stat.ML]


\bibitem[Mehrotra et~al\mbox{.}(2019)]%
        {mehrotra2019jointly}
\bibfield{author}{\bibinfo{person}{Rishabh Mehrotra}, \bibinfo{person}{Mounia
  Lalmas}, \bibinfo{person}{Doug Kenney}, \bibinfo{person}{Thomas Lim-Meng},
  {and} \bibinfo{person}{Golli Hashemian}.} \bibinfo{year}{2019}\natexlab{}.
\newblock \showarticletitle{Jointly leveraging intent and interaction signals
  to predict user satisfaction with slate recommendations}. In
  \bibinfo{booktitle}{\emph{The World Wide Web Conference}}.
  \bibinfo{pages}{1256--1267}.
\newblock


\bibitem[Pancha et~al\mbox{.}(2022)]%
        {pancha2022pinnerformer}
\bibfield{author}{\bibinfo{person}{Nikil Pancha}, \bibinfo{person}{Andrew
  Zhai}, \bibinfo{person}{Jure Leskovec}, {and} \bibinfo{person}{Charles
  Rosenberg}.} \bibinfo{year}{2022}\natexlab{}.
\newblock \showarticletitle{PinnerFormer: Sequence Modeling for User
  Representation at Pinterest}.
\newblock \bibinfo{journal}{\emph{arXiv preprint arXiv:2205.04507}}
  (\bibinfo{year}{2022}).
\newblock


\bibitem[Papineni et~al\mbox{.}(2002)]%
        {papineni2002bleu}
\bibfield{author}{\bibinfo{person}{Kishore Papineni}, \bibinfo{person}{Salim
  Roukos}, \bibinfo{person}{Todd Ward}, {and} \bibinfo{person}{Wei-Jing Zhu}.}
  \bibinfo{year}{2002}\natexlab{}.
\newblock \showarticletitle{BLEU: A Method for Automatic Evaluation of Machine
  Translation}. \bibinfo{publisher}{Association for Computational Linguistics},
  \bibinfo{address}{USA}.
\newblock


\bibitem[Pi et~al\mbox{.}(2019)]%
        {pi2019practice}
\bibfield{author}{\bibinfo{person}{Qi Pi}, \bibinfo{person}{Weijie Bian},
  \bibinfo{person}{Guorui Zhou}, \bibinfo{person}{Xiaoqiang Zhu}, {and}
  \bibinfo{person}{Kun Gai}.} \bibinfo{year}{2019}\natexlab{}.
\newblock \showarticletitle{Practice on long sequential user behavior modeling
  for click-through rate prediction}. In \bibinfo{booktitle}{\emph{Proceedings
  of the 25th ACM SIGKDD International Conference on Knowledge Discovery \&
  Data Mining}}. \bibinfo{pages}{2671--2679}.
\newblock


\bibitem[Radlinski et~al\mbox{.}(2022)]%
        {radlinski2022natural}
\bibfield{author}{\bibinfo{person}{Filip Radlinski}, \bibinfo{person}{Krisztian
  Balog}, \bibinfo{person}{Fernando Diaz}, \bibinfo{person}{Lucas Dixon}, {and}
  \bibinfo{person}{Ben Wedin}.} \bibinfo{year}{2022}\natexlab{}.
\newblock \showarticletitle{On Natural Language User Profiles for Transparent
  and Scrutable Recommendation}.
\newblock \bibinfo{journal}{\emph{arXiv preprint arXiv:2205.09403}}
  (\bibinfo{year}{2022}).
\newblock


\bibitem[Radlinski et~al\mbox{.}(2010)]%
        {radlinski2010inferring}
\bibfield{author}{\bibinfo{person}{Filip Radlinski}, \bibinfo{person}{Martin
  Szummer}, {and} \bibinfo{person}{Nick Craswell}.}
  \bibinfo{year}{2010}\natexlab{}.
\newblock \showarticletitle{Inferring query intent from reformulations and
  clicks}. In \bibinfo{booktitle}{\emph{Proceedings of the 19th international
  conference on World wide web}}. \bibinfo{pages}{1171--1172}.
\newblock


\bibitem[Rose and Levinson(2004)]%
        {rose2004understanding}
\bibfield{author}{\bibinfo{person}{Daniel~E Rose} {and} \bibinfo{person}{Danny
  Levinson}.} \bibinfo{year}{2004}\natexlab{}.
\newblock \showarticletitle{Understanding user goals in web search}. In
  \bibinfo{booktitle}{\emph{Proceedings of the 13th international conference on
  World Wide Web}}. \bibinfo{pages}{13--19}.
\newblock


\bibitem[Safavi et~al\mbox{.}(2020)]%
        {safavi2020toward}
\bibfield{author}{\bibinfo{person}{Tara Safavi}, \bibinfo{person}{Adam
  Fourney}, \bibinfo{person}{Robert Sim}, \bibinfo{person}{Marcin Juraszek},
  \bibinfo{person}{Shane Williams}, \bibinfo{person}{Ned Friend},
  \bibinfo{person}{Danai Koutra}, {and} \bibinfo{person}{Paul~N Bennett}.}
  \bibinfo{year}{2020}\natexlab{}.
\newblock \showarticletitle{Toward activity discovery in the personal web}. In
  \bibinfo{booktitle}{\emph{Proceedings of the 13th International Conference on
  Web Search and Data Mining}}. \bibinfo{pages}{492--500}.
\newblock


\bibitem[Sarwar et~al\mbox{.}(2001)]%
        {sarwar2001item}
\bibfield{author}{\bibinfo{person}{Badrul Sarwar}, \bibinfo{person}{George
  Karypis}, \bibinfo{person}{Joseph Konstan}, {and} \bibinfo{person}{John
  Riedl}.} \bibinfo{year}{2001}\natexlab{}.
\newblock \showarticletitle{Item-based collaborative filtering recommendation
  algorithms}. In \bibinfo{booktitle}{\emph{Proceedings of the 10th
  international conference on World Wide Web}}. \bibinfo{pages}{285--295}.
\newblock


\bibitem[Scaiella et~al\mbox{.}(2012)]%
        {37745}
\bibfield{author}{\bibinfo{person}{Ugo Scaiella}, \bibinfo{person}{Paolo
  Ferragina}, \bibinfo{person}{Andrea Marino}, {and}
  \bibinfo{person}{Massimiliano Ciaramita}.} \bibinfo{year}{2012}\natexlab{}.
\newblock \showarticletitle{Topical clustering of search results}. In
  \bibinfo{booktitle}{\emph{Proceedings of the fifth ACM international
  conference on Web search and data mining}}. \bibinfo{address}{New York, NY,
  USA}, \bibinfo{pages}{223--232}.
\newblock


\bibitem[Sellam et~al\mbox{.}(2020)]%
        {sellam2020bleurt}
\bibfield{author}{\bibinfo{person}{Thibault Sellam}, \bibinfo{person}{Dipanjan
  Das}, {and} \bibinfo{person}{Ankur~P. Parikh}.}
  \bibinfo{year}{2020}\natexlab{}.
\newblock \bibinfo{title}{BLEURT: Learning Robust Metrics for Text Generation}.
\newblock
\newblock
\showeprint[arxiv]{2004.04696}~[cs.CL]


\bibitem[Shahaf and Guestrin(2010)]%
        {shahaf2010connecting}
\bibfield{author}{\bibinfo{person}{Dafna Shahaf} {and} \bibinfo{person}{Carlos
  Guestrin}.} \bibinfo{year}{2010}\natexlab{}.
\newblock \showarticletitle{Connecting the dots between news articles}. In
  \bibinfo{booktitle}{\emph{Proceedings of the 16th ACM SIGKDD international
  conference on Knowledge discovery and data mining}}.
  \bibinfo{pages}{623--632}.
\newblock


\bibitem[Shahaf et~al\mbox{.}(2012)]%
        {shahaf2012trains}
\bibfield{author}{\bibinfo{person}{Dafna Shahaf}, \bibinfo{person}{Carlos
  Guestrin}, {and} \bibinfo{person}{Eric Horvitz}.}
  \bibinfo{year}{2012}\natexlab{}.
\newblock \showarticletitle{Trains of thought: Generating information maps}. In
  \bibinfo{booktitle}{\emph{Proceedings of the 21st international conference on
  World Wide Web}}. \bibinfo{pages}{899--908}.
\newblock


\bibitem[Song and Guo(2016)]%
        {song2016query}
\bibfield{author}{\bibinfo{person}{Yang Song} {and} \bibinfo{person}{Qi Guo}.}
  \bibinfo{year}{2016}\natexlab{}.
\newblock \showarticletitle{Query-less: Predicting task repetition for nextgen
  proactive search and recommendation engines}. In
  \bibinfo{booktitle}{\emph{Proceedings of the 25th International Conference on
  World Wide Web}}. \bibinfo{pages}{543--553}.
\newblock


\bibitem[Srivastava(2022)]%
        {srivastava2022imitation}
\bibfield{author}{\bibinfo{person}{Aarohi et~al. Srivastava}.}
  \bibinfo{year}{2022}\natexlab{}.
\newblock \showarticletitle{Beyond the imitation game: Quantifying and
  extrapolating the capabilities of language models}.
\newblock \bibinfo{journal}{\emph{arXiv preprint arXiv:2206.04615}}
  (\bibinfo{year}{2022}).
\newblock


\bibitem[Sun et~al\mbox{.}(2019)]%
        {sun2019bert4rec}
\bibfield{author}{\bibinfo{person}{Fei Sun}, \bibinfo{person}{Jun Liu},
  \bibinfo{person}{Jian Wu}, \bibinfo{person}{Changhua Pei},
  \bibinfo{person}{Xiao Lin}, \bibinfo{person}{Wenwu Ou}, {and}
  \bibinfo{person}{Peng Jiang}.} \bibinfo{year}{2019}\natexlab{}.
\newblock \showarticletitle{BERT4Rec: Sequential recommendation with
  bidirectional encoder representations from transformer}. In
  \bibinfo{booktitle}{\emph{Proceedings of the 28th ACM international
  conference on information and knowledge management}}.
  \bibinfo{pages}{1441--1450}.
\newblock


\bibitem[Sun et~al\mbox{.}(2016)]%
        {sun2016contextual}
\bibfield{author}{\bibinfo{person}{Yu Sun}, \bibinfo{person}{Nicholas~Jing
  Yuan}, \bibinfo{person}{Yingzi Wang}, \bibinfo{person}{Xing Xie},
  \bibinfo{person}{Kieran McDonald}, {and} \bibinfo{person}{Rui Zhang}.}
  \bibinfo{year}{2016}\natexlab{}.
\newblock \showarticletitle{Contextual intent tracking for personal
  assistants}. In \bibinfo{booktitle}{\emph{Proceedings of the 22nd ACM SIGKDD
  International Conference on Knowledge Discovery and Data Mining}}.
  \bibinfo{pages}{273--282}.
\newblock


\bibitem[Thoppilan et~al\mbox{.}(2022)]%
        {thoppilan2022lamda}
\bibfield{author}{\bibinfo{person}{Romal Thoppilan}, \bibinfo{person}{Daniel
  De~Freitas}, \bibinfo{person}{Jamie Hall}, \bibinfo{person}{Noam Shazeer},
  \bibinfo{person}{Apoorv Kulshreshtha}, \bibinfo{person}{Heng-Tze Cheng},
  \bibinfo{person}{Alicia Jin}, \bibinfo{person}{Taylor Bos},
  \bibinfo{person}{Leslie Baker}, \bibinfo{person}{Yu Du}, {et~al\mbox{.}}}
  \bibinfo{year}{2022}\natexlab{}.
\newblock \showarticletitle{Lamda: Language models for dialog applications}.
\newblock \bibinfo{journal}{\emph{arXiv preprint arXiv:2201.08239}}
  (\bibinfo{year}{2022}).
\newblock


\bibitem[Tintarev(2007)]%
        {tintarev2007explanations}
\bibfield{author}{\bibinfo{person}{Nava Tintarev}.}
  \bibinfo{year}{2007}\natexlab{}.
\newblock \showarticletitle{Explanations of recommendations}. In
  \bibinfo{booktitle}{\emph{Proceedings of the 2007 ACM conference on
  Recommender systems}}. \bibinfo{pages}{203--206}.
\newblock


\bibitem[Wang et~al\mbox{.}(2020)]%
        {wang2020superglue}
\bibfield{author}{\bibinfo{person}{Alex Wang}, \bibinfo{person}{Yada
  Pruksachatkun}, \bibinfo{person}{Nikita Nangia}, \bibinfo{person}{Amanpreet
  Singh}, \bibinfo{person}{Julian Michael}, \bibinfo{person}{Felix Hill},
  \bibinfo{person}{Omer Levy}, {and} \bibinfo{person}{Samuel~R. Bowman}.}
  \bibinfo{year}{2020}\natexlab{}.
\newblock \bibinfo{title}{SuperGLUE: A Stickier Benchmark for General-Purpose
  Language Understanding Systems}.
\newblock
\newblock
\showeprint[arxiv]{1905.00537}~[cs.CL]


\bibitem[Wei et~al\mbox{.}(2022)]%
        {wei2022finetuned}
\bibfield{author}{\bibinfo{person}{Jason Wei}, \bibinfo{person}{Maarten Bosma},
  \bibinfo{person}{Vincent~Y. Zhao}, \bibinfo{person}{Kelvin Guu},
  \bibinfo{person}{Adams~Wei Yu}, \bibinfo{person}{Brian Lester},
  \bibinfo{person}{Nan Du}, \bibinfo{person}{Andrew~M. Dai}, {and}
  \bibinfo{person}{Quoc~V. Le}.} \bibinfo{year}{2022}\natexlab{}.
\newblock \bibinfo{title}{Finetuned Language Models Are Zero-Shot Learners}.
\newblock
\newblock
\showeprint[arxiv]{2109.01652}~[cs.CL]


\bibitem[White et~al\mbox{.}(2013)]%
        {white2013enhancing}
\bibfield{author}{\bibinfo{person}{Ryen~W White}, \bibinfo{person}{Wei Chu},
  \bibinfo{person}{Ahmed Hassan}, \bibinfo{person}{Xiaodong He},
  \bibinfo{person}{Yang Song}, {and} \bibinfo{person}{Hongning Wang}.}
  \bibinfo{year}{2013}\natexlab{}.
\newblock \showarticletitle{Enhancing personalized search by mining and
  modeling task behavior}. In \bibinfo{booktitle}{\emph{Proceedings of the 22nd
  international conference on World Wide Web}}. \bibinfo{pages}{1411--1420}.
\newblock


\bibitem[Wu et~al\mbox{.}(2017)]%
        {wu2017recurrent}
\bibfield{author}{\bibinfo{person}{Chao-Yuan Wu}, \bibinfo{person}{Amr Ahmed},
  \bibinfo{person}{Alex Beutel}, \bibinfo{person}{Alexander~J Smola}, {and}
  \bibinfo{person}{How Jing}.} \bibinfo{year}{2017}\natexlab{}.
\newblock \showarticletitle{Recurrent recommender networks}. In
  \bibinfo{booktitle}{\emph{Proceedings of the tenth ACM international
  conference on web search and data mining}}. \bibinfo{pages}{495--503}.
\newblock


\bibitem[Yin and Shah(2010)]%
        {yin2010building}
\bibfield{author}{\bibinfo{person}{Xiaoxin Yin} {and} \bibinfo{person}{Sarthak
  Shah}.} \bibinfo{year}{2010}\natexlab{}.
\newblock \showarticletitle{Building taxonomy of web search intents for name
  entity queries}. In \bibinfo{booktitle}{\emph{Proceedings of the 19th
  international conference on World wide web}}. \bibinfo{pages}{1001--1010}.
\newblock


\bibitem[Zaheer et~al\mbox{.}(2017)]%
        {zaheer2017latent}
\bibfield{author}{\bibinfo{person}{Manzil Zaheer}, \bibinfo{person}{Amr Ahmed},
  {and} \bibinfo{person}{Alexander~J Smola}.} \bibinfo{year}{2017}\natexlab{}.
\newblock \showarticletitle{Latent LSTM allocation: Joint clustering and
  non-linear dynamic modeling of sequence data}. In
  \bibinfo{booktitle}{\emph{International Conference on Machine Learning}}.
  PMLR, \bibinfo{pages}{3967--3976}.
\newblock


\bibitem[Zhang et~al\mbox{.}(2011)]%
        {zhang2011user}
\bibfield{author}{\bibinfo{person}{Yuchen Zhang}, \bibinfo{person}{Weizhu
  Chen}, \bibinfo{person}{Dong Wang}, {and} \bibinfo{person}{Qiang Yang}.}
  \bibinfo{year}{2011}\natexlab{}.
\newblock \showarticletitle{User-click modeling for understanding and
  predicting search-behavior}. In \bibinfo{booktitle}{\emph{Proceedings of the
  17th ACM SIGKDD international conference on Knowledge discovery and data
  mining}}. \bibinfo{pages}{1388--1396}.
\newblock


\bibitem[Zhou et~al\mbox{.}(2019)]%
        {zhou2019deep}
\bibfield{author}{\bibinfo{person}{Guorui Zhou}, \bibinfo{person}{Na Mou},
  \bibinfo{person}{Ying Fan}, \bibinfo{person}{Qi Pi}, \bibinfo{person}{Weijie
  Bian}, \bibinfo{person}{Chang Zhou}, \bibinfo{person}{Xiaoqiang Zhu}, {and}
  \bibinfo{person}{Kun Gai}.} \bibinfo{year}{2019}\natexlab{}.
\newblock \showarticletitle{Deep interest evolution network for click-through
  rate prediction}. In \bibinfo{booktitle}{\emph{Proceedings of the AAAI
  conference on artificial intelligence}}, Vol.~\bibinfo{volume}{33}.
  \bibinfo{pages}{5941--5948}.
\newblock


\bibitem[Ziegler et~al\mbox{.}(2020)]%
        {ziegler2020finetuning}
\bibfield{author}{\bibinfo{person}{Daniel~M. Ziegler}, \bibinfo{person}{Nisan
  Stiennon}, \bibinfo{person}{Jeffrey Wu}, \bibinfo{person}{Tom~B. Brown},
  \bibinfo{person}{Alec Radford}, \bibinfo{person}{Dario Amodei},
  \bibinfo{person}{Paul Christiano}, {and} \bibinfo{person}{Geoffrey Irving}.}
  \bibinfo{year}{2020}\natexlab{}.
\newblock \bibinfo{title}{Fine-Tuning Language Models from Human Preferences}.
\newblock
\newblock
\showeprint[arxiv]{1909.08593}~[cs.CL]


\end{thebibliography}
